\documentclass[two column]{article}

\usepackage[round]{natbib}

\usepackage{xcolor}

\usepackage[utf8]{inputenc} 
\usepackage[T1]{fontenc}    
\usepackage{hyperref}       
\usepackage{url}            
\usepackage{booktabs}       
\usepackage{amsfonts}       
\usepackage{amsmath}
\usepackage{mathtools}
\usepackage{nicefrac}       
\usepackage{microtype}      
\usepackage{xcolor}         
\usepackage{amsthm}
\usepackage{natbib}
\usepackage{algorithm}
\usepackage{algpseudocode}
\usepackage{array} 
\usepackage{ulem}
\usepackage{caption}
\usepackage{subcaption}
\usepackage{float}

\usepackage{multirow}
\newcommand{\mb}[1]{\mathbf{#1}}
\newcommand{\bs}[1]{\boldsymbol{#1}}
\newcommand{\mc}[1]{\mathcal{#1}}

\usepackage{siunitx}
\usepackage[margin=0.7in]{geometry}
\usepackage{authblk}

\newcommand{\imagepath}{./figures/innowell/v3}

\usepackage[compact]{titlesec}

\titlespacing*{\section}
{0pt}{*1.5}{*1}

\titlespacing*{\subsection}
{0pt}{*1.25}{*0.75}

\titlespacing*{\subsubsection}
{0pt}{*1}{*0.5}

\title{Covariate Dependent Mixture of Bayesian Networks}

\author[1]{Roman Marchant\footnote{Corresponding author: roman.marchant@uts.edu.au}}
\author[2]{Dario Draca}
\author[1]{Gilad Francis}
\author[1]{Sahand Assadzadeh}
\author[3]{Mathew Varidel}
\author[3]{Frank Iorfino}
\author[1]{Sally Cripps}

\affil[1]{Human Technology Institute, The University of Technology Sydney, Australia}
\affil[2]{School of Mathematics and Statistics, The University of Sydney, Australia}
\affil[3]{Brain and Mind Centre, The University of Sydney, Australia}

\date{January 2025}

\begin{document}

%

%

\twocolumn[

\maketitle

]

\begin{abstract}
Learning the structure of Bayesian networks from data provides insights into underlying processes and the causal relationships that generate the data, but its usefulness depends on the homogeneity of the data population—a condition often violated in real-world applications. In such cases, using a single network structure for inference can be misleading, as it may not capture sub-population differences. To address this, we propose a novel approach of modelling a mixture of Bayesian networks where component probabilities depend on individual characteristics. Our method identifies both network structures and demographic predictors of sub-population membership, aiding personalised interventions. We evaluate our method through simulations and a youth mental health case study, demonstrating its potential to improve tailored interventions in health, education, and social policy.
\end{abstract}

\begingroup
\renewcommand\thefootnote{\text{*}}
\footnotetext{Corresponding author: roman.marchant@uts.edu.au}
\addtocounter{footnote}{0}
\endgroup

\section{Introduction}
\label{sec:intro}

Probabilistic graphical models, particularly {\it Bayesian Networks} (BNs), are powerful tools for modeling statistical dependencies over collections of random variables \citep{koller2009}. By encoding a set conditional independencies through a {\it Directed Acyclic Graph} (DAG), these models provide a succinct and intuitive representations of interactions among locally connected variables. Although conditional independence structures do not uniquely define the causal relationships \citep{dawid2010}, they do provide a set of possible causal models that are captured within the same Markov equivalence class. These equivalence classes can be represented by a {\it Completed Partially Directed Acyclic Graph} (CPDAG) \citep{dawid2010}. Distinguishing between different causal models within an equivalence class typically requires additional assumptions encoded in prior knowledge or interventional data \citep{Pearl2009}. For an excellent survey and overview of these topics, see \citet{dawid2024}.

Learning the structure of a BN from data is complex; the cardinality of the discrete network search-space grows in a super-exponential fashion with the number of variables, making an exact search implausible outside of moderately sized domains containing 10 - 15 variables. Numerous methods for structure learning have been developed; a recent overview is presented by \citet{vowels_dya_2023}.

However the validity of inference tasks is often less dependent on the algorithm used to estimate the network structure and accompanying posterior distribution, and more dependent on the appropriateness of assumptions underpinning the {\it Data Generating Process} (DGP) of the network. One example of this, which has been the subject of much research over the years, is the presence of unobserved confounders, potentially giving rise to spurious edges in learned network \citep{Chobtham2020,Chobtham2022,haggstrom2018,Higashigaki2010}.

The assumption of the homogeneity of the DGP, has started to receive attention, with some methods using mixture models to relax the assumption of homogeneity in the DGP,  (\cite{saeed2020,stobl2023,varıcı2024}). However, these methods use constraint-based optimization techniques to obtain only point estimates of component graph structures and probabilities.  To our knowledge, with the notable exception of \citet{Castelletti2020},  there are no methods that model the joint posterior probabilities of the graph structure components and the corresponding component probabilities, which is essential for uncertainty quantification and decision-making.   Furthermore, the mixture models (\cite{saeed2020,stobl2023,varıcı2024}) do not parametrise the mixing probabilities to depend upon covariates. This is unfortunate, as population heterogeneity is likely to arise from particular subsections of the population, where demographic factors such as an individual’s age, gender, and socio-economic status can directly or indirectly affect a wide range of life outcomes. Learning a mixture network without uncertainty quantification or without the ability to specifically encode potential drivers of the heterogeneity scenarios could lead to sub-optimal decision by practitioners who use such models.

The purpose of this paper is to propose a fully probabilistic, yet computationally tractable, methodology for estimating and inferring multiple graph structures which result from heterogeneous DGPs. In developing our model we are motivated by two factors. The first is to improve the explainability and interpretability of model outputs by making inference robust to the assumption of a homogeneous DGP. The second is to reduce the computational complexity of the structure learning process for large networks. To these ends we propose that the data are generated from a finite, but unknown number of mixture components where the probability that an individual's graph structure belongs to a mixture component depends upon an individual's features. This is a variant of a mixture-of-experts model \citep{ME1995}, which is receiving renewed interest to model complexity in a parsimonious manner \citep{chen2022, oldfield2024}, where the experts are BNs and the gating or mixing function is a multinomial logistic model. Importantly, our work differentiates from \citeauthor{Castelletti2020} (\citeyear{Castelletti2020}) since we divide the data into those features of an individual which have the potential for an intervention and those which do not. Those features which have the potential for an intervention we term {\it modifiables} while those that do not are termed {\it non-modifiables}. The former group are the features in the graph while the latter are the features in the mixing function. The motivation for this formulation is that by including features which cannot be subject to an intervention, such as various demographic factors, in the mixture weights, our model can better accommodate the latent class structure in our dataset, and allow for the identification of potential interventions without significantly impacting the computational complexity of the learning process.

The paper is structured as follows: Section \ref{sec:mixtures_of_bayesian_networks} formally defines our model. Section \ref{sec:inference_algorithm} describes the inference procedure used to obtain a sample estimate of the joint posterior of multiple graph structures, the corresponding parameters and the gating functions. In Section \ref{sec:experiments}, we evaluate the performance of our mixtures model and inference procedure by fitting both synthetic datasets with known ground truth DAGs, and a real world example involving youth mental health. Finally, Section \ref{sec:conclusion} provides conclusions, discusses the limitations of the methods presented and suggests directions for future work.

\section{Mixtures of Bayesian Networks, Model and Priors}
\label{sec:mixtures_of_bayesian_networks}
\subsection{Model}

Suppose we have variables $\bs y\in\mathbb{R}^M$ and $\bs x\in\mathbb{R}^P$, which correspond to $M$ modifiable and $P$ non-modifiable features of an individual, respectively. We model the distribution of $\bs y$ conditional on $\bs x$, namely $p(\bs y|\bs x)$ as arising from a finite but unknown number of $K$ mixture components.  The  components in the mixture are DAGs, whose structure, denoted by $\mc{G}$, depends solely on  $\mathbf{y}$. The probability of belonging to a mixture component, denoted by $\pi$,  depends solely on $\mathbf{x}$, so that
\begin{equation}
p(\bs y|\bs x)=\sum_{k=1}^K p(\bs y|\mathcal{G}_k)\pi_k(\bs x),
\label{eqn_mix}
\end{equation}
where the subscript $k$ indexes the components in the mixture for $k=1,\ldots, K$.
\subsection{Model for Mixture Components}
The $K$ components in \eqref{eqn_mix} are modelled as $M$ node BNs. Each of these BNs is defined by its structure $\mathcal{G}_k$ and associated parameters, $\tilde{\bs \theta}_k=(\bs\theta_{1,k},\ldots,\theta_{M,k})$. We assume that the distribution of the nodes in the graph, conditional on its parents are Gaussian and $\tilde{\bs \theta}_k$, contains the conditional means and variances of these nodes. To define the mean and variance of the $M$ nodes we introduce the following notation; let  $\gamma_{k,j,i}=1$ if $y_j$ is a parent of $y_i$ in graph structure $\mathcal{G}_k$ and define $P_{k,i}:=\{j; \gamma_{k,j,i}=1\}$ to be the set of indices which identify the parents of node $i$ in graph $\mc{G}_k$ for $i=1,\ldots,M$  and $k=1,\ldots K$. The distribution of each node is then
\begin{equation} \label{cond.like.eqn}
    p(y_i \; \vert \;  \mathcal{G}_k, \boldsymbol{\theta}_{k,i}) := \mathcal{N}(m_{k,i} + \boldsymbol{b}_{k,i} \cdot \bs{y}_{P_{k,i}}, v_{k,i}),
\end{equation}
so that $\bs\theta_{k,i}:= \{(m_{k,i}, \bs{b}_{k,i}, v_{k,i})\}_{1 \le i \le M}$.

It can be shown that this family of linear Gaussian regressions amounts to a joint distribution for $\boldsymbol{y}=(y_1,\ldots,y_m)$ which is multivariate normal with some mean $\boldsymbol{\mu_k} \in \mathbb{R}^{M}$ and precision $W_k \in \mathbb{R}^{M\times M}$ . The transformation from $(\mathcal{G}_k, \boldsymbol{\theta_k}) \rightarrow (\boldsymbol{\mu_k}, W_k)$ is provided by \citet{anderson2009}.

The graph structure $\mc G_k$ of a BN encodes the conditional independence structure of $\bs y$, such that each node $y_i$, conditional on its parents $\bs y_{P{k,i}}$, is independent of its non-descendants \citep{koller2009}.  This {\it Markov} property allows us to decompose the joint distribution into a  product of conditionally independent distributions,
\begin{equation}
 p(\bs y| \mathcal{G}_k, \tilde{\boldsymbol{\theta}}_{k}) =\prod_{i=1}^M p(y_i \; \vert \; \boldsymbol{y}_{P_{k,i}}, \boldsymbol{\theta}_{k,i})
\end{equation}
with $p(y_i \; \vert \; \boldsymbol{y}_{P_{k,i}}, \boldsymbol{\theta}_{k,i})$ given by \eqref{cond.like.eqn}.

\subsubsection{Model for mixture probabilities}
The mixture probabilities $\bs\pi=(\pi_1,\ldots,\pi_K)$, depend on $\bs x$ and parameters which we denote by $\tilde{\boldsymbol{\beta}} = (\boldsymbol{\beta}_{1}, ..., \boldsymbol{\beta}_{K})$. To define the model for the $\bs \pi$'s, we introduce latent indicator variables, 
\begin{equation} 
    \boldsymbol{z} := \{ z \in \{1, ..., K\} \}
\end{equation}

where $z=k$, if $\bs y$ is generated by the $k^{th}$ DGP, so that  $\pi_k(\bs x, \tilde{\boldsymbol{\beta}}) =\Pr(z=k|\bs x, \tilde{\boldsymbol{\beta}})$
is the probability that $\bs y$ has a distribution defined by $(\mc{G}_k,\tilde{\bs\theta}_k)$ for $k=1,\ldots,K$. 

This probability is modelled as a multinomial logistic regression, so that 
\begin{equation} \label{eq:weight.vector.eqn}
    \pi_k(\bs{x},\tilde{\boldsymbol{\beta}}) = \frac{\exp\left(g(\bs{x},\boldsymbol{\beta}_k)\right)}{\sum_{j=1}^K \exp\left(g(\bs{x}, \boldsymbol{\beta}_j)\right)}\,\,,
\end{equation}
with $k=1,\ldots,K$ and $\bs\beta_K=:0$ for identifiability. We take $g(\bs{x}, \boldsymbol{\beta}_j)=\bs{x}\cdot \boldsymbol{\beta}_j$ but note that more expressive choices for $g$ are possible.
\subsubsection{Likelihood}
Suppose that we have a dataset $\mathcal{D}$ containing $N$ independent observation vectors $(\bs{x}, \bs{y})$, with $\mc{D} := \{  \mc{Y},\mc{X}\}$ where $\mc Y:= \{\bs y^{(1)},\dots,\bs y^{(N)}\}$ and $\mc X:= \{\bs x^{(1)},\dots,\bs x^{(N)}\}$ and similarly define the set of latent indicator variables $\mc Z:= \{\bs z^{(1)},\dots,\bs z^{(N)}\}$. Let the collection of graphs structures and associated parameter vectors in a mixture of $K$ components be denoted by $\bs{G}:=\{\mathcal{G}_1,\ldots, \mathcal{G}_K\}$ and $\bs\Theta:=\{\tilde{\bs\theta}_1,\ldots,\tilde{\bs\theta}_K\}$ respectively.  Conditional on $\bs{x}^{(n)}$, $\bs{G}$, $\tilde{\bs{\beta}}$, and $\bs\Theta$, the realization $\bs{y}^{(n)}$ is assumed to be generated from the process
\begin{multline} \label{model.likelihood.eqn}
    p(\bs{y}^{(n)}\vert \bs{x}^{(n)},\bs{G},\bs{\Theta},\tilde{\bs{\beta}}) \\
    = \sum^{K}_{k=1} \pi_{k}(\bs{x}^{(n)}, \tilde{\boldsymbol{\beta}}) 
    p(\bs y^{(n)} \vert z^{(n)}=k, \mc{G}_k, \bs{\theta}_k)\,\,,
\end{multline}
where $p(\bs y^{(n)} \vert z^{(n)}=k, \mc{G}_k, \bs{\theta}_k)$ is given by \eqref{cond.like.eqn} and $\pi_{k}(\bs{x}^{(n)}, \tilde{\boldsymbol{\beta}})$ is given by \eqref{eq:weight.vector.eqn}.  Then, the likelihood of $\mc Y$ conditional on $\mc X$ is
\begin{equation}
    p(\mc{Y}\vert\mc{X},\bs G,\bs\Theta)=\prod_{n=1}^Np(\bs{y}^{(n)}| \bs{x}^{(n)},\bs{G},\bs{\Theta},\tilde{\bs{\beta}}).
\end{equation}

\subsection{Priors}\label{subsec:priors}
\subsubsection{ Priors on mixture components}
For the mixture components we place priors over $(\mc{G}_k,\tilde{\bs\theta}_k)$ as follows
\begin{equation}
   p(\mc{G}_k,\tilde{\bs\theta}_k)=\Pr(\mc{G}_k)p(\tilde{\bs\theta}_k|\mc{G}_k).
\end{equation}
The prior on $\mc{G}$ is discrete uniform with $\Pr(\mc{G}_k)=1/\mathcal{N}_G$, where $\mathcal{N}_G$ is the number of all possible graphs.  
We assume
\begin{equation}
p(\tilde{\boldsymbol{\theta}}_k \vert \mc G_k)=\prod_{i=1}^M p(\bs\theta_{k,i}\vert \mc G_k)\,\,,    
\end{equation}
where $\bs\theta_{k,i}=(m_{k,i}, \boldsymbol{b}_{k,i}, v_{k,i})$. We follow \cite{Geiger2002,geiger2021}, and take the prior $p(m_{k,i}, \bs{b}_{k,i}, v_{k,i}|\mc G_k)=p(m_{k,i}, \bs{b}_{k,i}| v_{k,i},\mc G_k)p(v_{i,k}|\mc G_k)$, where
\begin{align} \label{theta.prior}
    p(\boldsymbol{b}_{k,i}, m_{k,i} \vert v_{k,i}, \mc G_k)
    &\sim
    \mathcal{N}(\boldsymbol{0}, v_{k,i} \times I_M),\\
    \,\, v_{k,i} \ \vert \ \mc G_k &\sim \text{IG}\left(\frac{3 + ||P_{k,i}||}{2},\ \frac{1}{2}\right)
\end{align}
This prior leads to a closed-form expression for
\begin{equation}
p(\bs y|\mc G_k)=\int(p(\bs y|\mc G_k,\tilde{\bs\theta}_k)
p(\tilde{\bs\theta}_k|\mc G_k)d\tilde{\bs\theta}_k),
\end{equation}
with $\log\left(p(\bs y|\mc G_k)\right)$  known as the BGe score \citep{Geiger2002} and is available as part of an well-known R package \footnote{\url{https://cran.r-project.org/web/packages/BiDAG/index.html}}.

\subsubsection{Priors on mixing probabilities}
The prior for the latent class indicator vector, $\bs z$, is multinomial with probabilities that depend on $\bs{x}$ and the parameters $\tilde{\boldsymbol{\beta}}$. We assume
\begin{equation}
p(\tilde{\boldsymbol{\beta}}) =\prod_{k=1}^{K-1}p(\bs\beta_k)\,\,,    
\end{equation}
with $\bs\beta_k\sim \mc N(0,cI_{P+1})$ for $k=1,\ldots, K$.

The full model description is as follows,
\begin{align}
    \bs{y}^{(n)} \mid \boldsymbol{\Theta}, \bs{G}, z^{(n)} & \sim P(\bs{y}^{(n)} \mid \mc G_{z^{(n)}}, \tilde{\boldsymbol{\theta}}_{z^{(n)}}),
    \label{eq:model_y} \\
    z^{(n)} \mid \boldsymbol{x}^{(n)}, \tilde{\boldsymbol{\beta}} & \sim \text{MN}(1, \boldsymbol{\pi}(\boldsymbol{x}^{(n)}; \tilde{\boldsymbol{\beta}})),  \label{eq:model_z} \\
    (\boldsymbol{b}_{k,i}, m_{k,i}) \mid v_{k,i}, \mc G_k & \sim \mathcal{N}(\boldsymbol{0}, v_{k,i} \times I_M), \label{eq:model_b_m} \\
    v_{k,i} \mid \mc G_k & \sim \text{IG}\left(\frac{3 + ||P_{k,i}|| }{2}, \frac{1}{2}\right), \label{eq:model_v} \\
    \mc G_k & \sim U(\mathcal{N}_G), \label{eq:model_G} \\
    \boldsymbol{\beta}_k & \sim \mathcal{N}(\boldsymbol{0}, c I_{P+1}), \label{eq:model_beta}
\end{align}
for $1 \le n \le N$, $1 \le i \le M$, $1 \le k \le K$, and MN denoting the Multinomial distribution.

\section{Inference of mixtures of Bayesian networks}
\label{sec:inference_algorithm}

The joint posterior distribution of all parameters is given by
\begin{multline} \label{joint.posterior}
    p(\mc Z, \bs G, \tilde{\boldsymbol{\beta}} \ \vert \ \mathcal{D}) \propto 
    \Bigg[\prod^{N}_{n=1} \exp{(\boldsymbol{x}^{(n)} \cdot \boldsymbol{\beta}_{z^{(n)}})} p(\boldsymbol{y}^{(n)} \vert \mc G_{z^{(n)}}) \Bigg]\\ \cdot \Bigg[ \prod^{K}_{k=1} p(\boldsymbol{\beta}_k) \Bigg] \,\,.
\end{multline}
Draws from the joint posterior $p(\mc Z, \bs G, \tilde{\boldsymbol{\beta}} \ \vert \ \mathcal{D})$ are obtained using {\it Markov chain Monte Carlo} (MCMC). The hierarchical structure of the model presented in \eqref{eq:model_y} to \eqref{eq:model_beta}, lends itself to a block Gibbs sampling scheme, detailed in Algorithm~\ref{alg:mcmc}. First, the latent indicator variables, $\mc Z$, are drawn; then the parameters of the mixing probabilities, $\tilde{\bs\beta}$, using the data augmentation scheme of \cite{Polson2012}; and finally, conditional on $\mc Z$, the $K$ individual DAG structures, $\bs G$, are drawn using {\it Partition MCMC} (PMCMC) \citep{kuipers2017}.

\begin{figure*}[t]
\centering
\begin{minipage}{\textwidth}
\begin{algorithm}[H]
\caption{Sampling from the Joint Posterior Distribution}
\label{alg:mcmc}
\begin{algorithmic}[1]
\State \textbf{Input:} Data $\mathcal{D}$, initial parameters $\mc{Z}^{[0]}$, $\tilde{\boldsymbol{\beta}}^{[0]}$, $\boldsymbol{G}^{[0]}$, number of iterations $T$
\State \textbf{Output:} Samples from $p(\mc{Z}, \boldsymbol{G},  \tilde{\boldsymbol{\beta}} \mid \mathcal{D})$

\State Initialize parameters $\mc{Z}^{[0]}$, $\tilde{\boldsymbol{\beta}}^{[0]}$,  $\bs{G}^{[0]}$
\For{$t = 1$ to $T$} \Comment{Main loop for MCMC iterations}
    \For{$n = 1$ to $N$} \Comment{For each individual}
        \State Draw $z^{(n)[t]}$ from $p(z^{(n)} \mid  \boldsymbol{G}^{[t-1]}, \tilde{\boldsymbol{\beta}}^{[t-1]}, \mathcal{D})$ \Comment{See Eq. \ref{eq:model_z} and \ref{eq:weight.vector.eqn}}
    \EndFor
    \For{$k = 1$ to $K$} \Comment{For each mixture component, update weights and network}
        \State Draw $\boldsymbol{\beta}_k^{[t]}$ from $p(\boldsymbol{\beta}_k^{[t]}|\mc Z^{[t]})$\Comment{See supplementary material}
        \State Draw $\mc{G}_k^{[t]}$ from $p(\mc{G}_k|\mc D,\mc Z^{[t]})$ using PMCMC \Comment{\citet{kuipers2017}}
       
    \EndFor
\EndFor
\State \textbf{return} $\{\boldsymbol{Z}^{[t]}, \tilde{\boldsymbol{\beta}}^{[t]}, \boldsymbol{G}^{[t]}\}_{t=1}^T$
\end{algorithmic}
\end{algorithm}
\end{minipage}
\end{figure*}

The individual expressions for each conditional posterior of $\mc{Z}$, $\tilde{\boldsymbol{\beta}}$, and $\boldsymbol{G}$ are presented in the Supplementary Material.

\section{Experiments}
\label{sec:experiments}

In this section we first examine the behaviour of our sampling scheme on synthetic data generated from a known mixture of Gaussian BNs, where we can validate our method and quantify goodness of fit based over ground truth data. We compare the performance of our method,  which allows for heterogeneity with the base case which assumes a homogeneous data generating process. Our results show that when the data generating process is homogeneous then our method performs as well as the base case, while when  data are generated from a heterogeneous process our method outperforms the base case method. 

We later show how our model and inference method provides superior explanatory capabilities for heterogeneous populations in a real-world example based on youth mental health.

\subsection{Synthetic Data Experiments}\label{subsec:synth_data} 

First, we assume that the number of mixture components, $K$, is known, and consider the similarity of our posterior network samples to their ground truth counterparts. Second, we examine a technique for choosing $K$ when this parameter is unknown.

The datasets used to assess structure recovery and mixture enumeration were generated from various random mixtures of Gaussian BN models, where the process used to generate the model and subsequent dataset is described by a combination $(\mc K, N, S, \mc C)$ of the following conditions: (1) number of true mixture components: $\mc K \in \{1, 2, 3, 4\}$; (2) number of observations per mixture component: $N \in \{100, 300, 500\}$; (3) network sparsity: $S \in \{\mbox{low},\mbox{high}\}$; (4) cluster density in non-modifiable space $\mc C \in \{\text{Sparse}, \text{Mid}, \text{Dense} \}$.

For every source model, each BN mixture component contained five nodes (modifiable features, $M=5$), where the conditional distribution for each node $i$ is parameterized by the triple $(m_i, \boldsymbol{b}_i, v_i)$. The elements of the regression coefficients $(m_i, \boldsymbol{b}_i)$ are restricted to $(-2,-1) \ \cup \ (1, 2)$, while the conditional variance $v_i$ lies within $(0.75,1.25)$. The structures of the source networks were obtained using the Erdős–Rényi model $G(M = 5,p) $ for generating random networks \citep{Erdos1959}. Here, $p$ refers to the probability that each edge is present in the graph, independently of others. The network sparsity parameter $S = \{\mbox{Low}, \mbox{High}\}$ correspond to $p=0.25$ and $p=0.5$ respectively (for all networks).

The most commonly used metric for quantifying the disparity between two networks is the {\it Structural Hamming Distance} (SHD) \citep{Tsamardinos2006}. For two networks $G$ and $G'$, the SHD equals the total number of edge modifications (i.e., additions, removals, and reversals) required to convert the CPDAG of $G$ to that of $G'$. However, because we are working with posterior samples of networks, we report the mean SHD between sampled networks and their corresponding ground truth structures. As there is ambiguity regarding the correct ground-truth structure with which to associate each of the $K$ posterior network structure samples, we compute the sum of the mean SHD across mixture components for every possible labelling, and report the minimal choice (hereafter referred to as the model SHD or MSHD).

\begin{figure}[t]
    \centering
    \begin{subfigure}[b]{0.23\textwidth}
        \centering
        \includegraphics[scale=0.26]{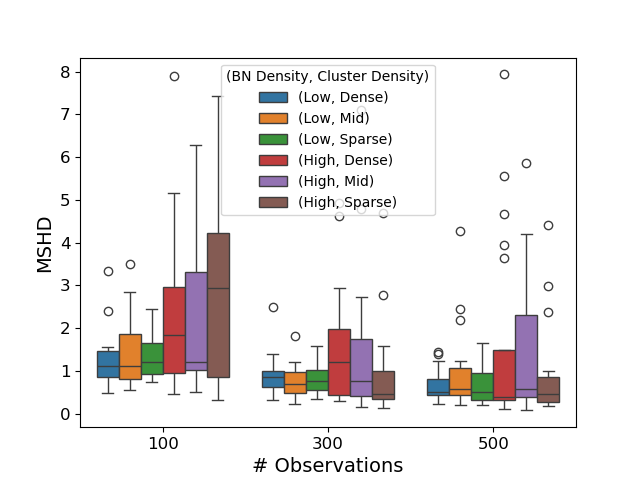}
        \caption{$\mc K = 1$}
    \end{subfigure}
    \begin{subfigure}[b]{0.23\textwidth}
        \centering
        \includegraphics[scale=0.26]{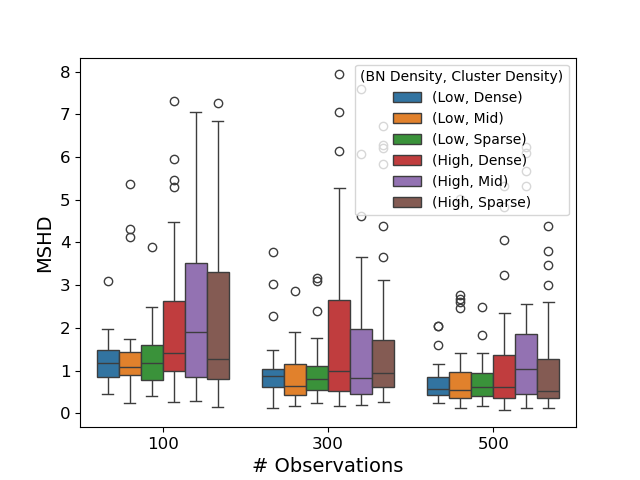}
        \caption{$\mc K = 2$}
    \end{subfigure}
    \hfill
    \begin{subfigure}[b]{0.23\textwidth}
        \centering
        \includegraphics[scale=0.26]{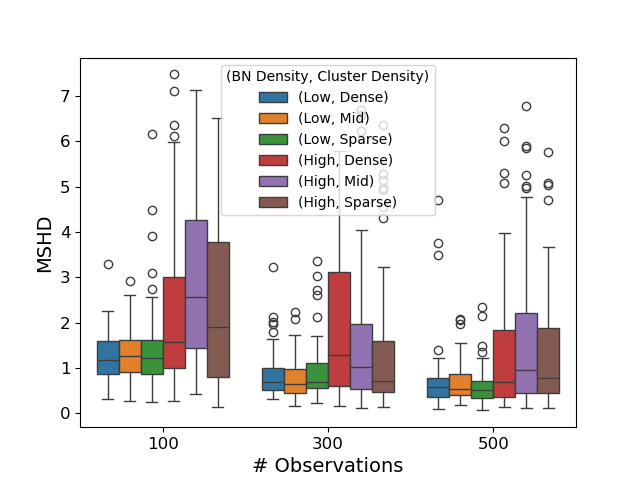}
        \caption{$\mc K = 3$}
    \end{subfigure}
    \begin{subfigure}[b]{0.23\textwidth}
        \centering
        \includegraphics[scale=0.26]{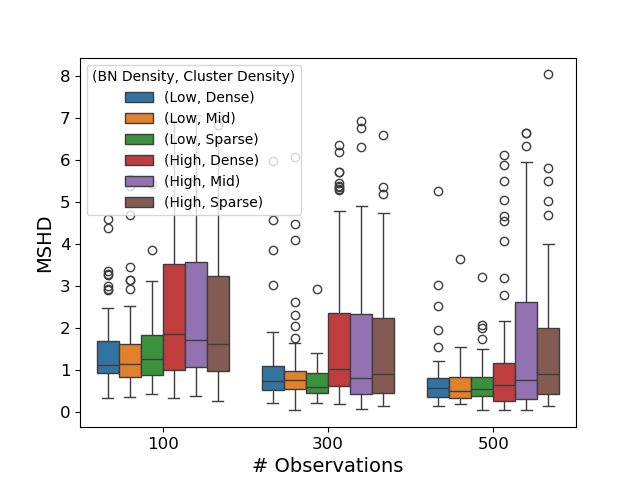}
        \caption{$\mc K = 4$}
    \end{subfigure}
    \caption{MSHD between estimated graphs and ground truth, for varying number of true mixture components, $\mc K$ as sub-figures, number of observations per mixture,$N$, as the horizontal axis; network sparsity $S$, cluster density in non-modifiables space $\mc C$ as the combinations in the legend.}
    \label{fig:MSHD}
\end{figure}

Figure \ref{fig:MSHD} depicts the average MSHD associated with every combination of conditions $(\mc K, n, S)$. Figure \ref{fig:MSHD} illustrates the following three points. First, the MSHD is below one, indicating that when $K = \mc K$, our approach converges to the true mixture of data-generating BNs. Second, the density of the clusters (the non-modifiable factors), does not impact the ability of the technique to uncover the true data generating BN. And third, and in contrast, the performance of the technique is impacted by the density of the BN, with more dense mixtures of networks being more difficult to estimate than less dense ones.

\subsubsection{Selection of Number of Mixtures}
\label{selection.criteria}

In many real-world problems involving mixture models, choosing the appropriate number of components $K$ is a well-known and challenging problem. There are numerous established solutions, such as information criteria, which are based on a suitably-penalized predictive likelihood. Alternatively, more computationally-intensive Bayesian methods, such as those involving reversible-jump algorithms \citep{green95}, can treat the number of mixture components as a model parameter, and yield posterior estimates of it.

We consider the selection of the number of mixtures, $K$, by utilising a criterion involving a cross-validation-based estimate of the number of mixture components. The use of cross-validation for model-selection and hyper-parameter tuning is well-established in the statistical machine learning literature. The selection criteria that we employ is a variant of this technique, which accommodates our Bayesian approach to parameter estimation. Thus, suppose we have a dataset divided into a training and test set, denoted by $\mc D$ and $\mc D^*=(\mc Y^*,\mc X^*)$ respectively.

We compute the {\it Log Marginal Posterior Predictive Density} (LMPPD) for the test data $p(\mc{Y}^*\vert\mc{X^*},\mc D)$ using the posterior samples to perform the required marginalization.

For each $1 \le \mc K \le 4$ we generate a model $M$ with mixed network sparsity, according to the procedure defined in section \ref{selection.criteria}. From each model, we sample a dataset $D$ with $200$ observations per mixture component. For each such dataset, we obtain estimates of LMPPD for models with $1 \le K \le 5$, using the procedure described in section \ref{selection.criteria}. Each of the $10 \times 4 \times 5$ runs of MCMC were carried out for $1500$ iterations using an i7 CPU with a total compute of $2 [kSU]$. 

\begin{figure}[t]
    \centering
    \begin{subfigure}[b]{0.23\textwidth}
        \centering
        \includegraphics[scale=0.26]{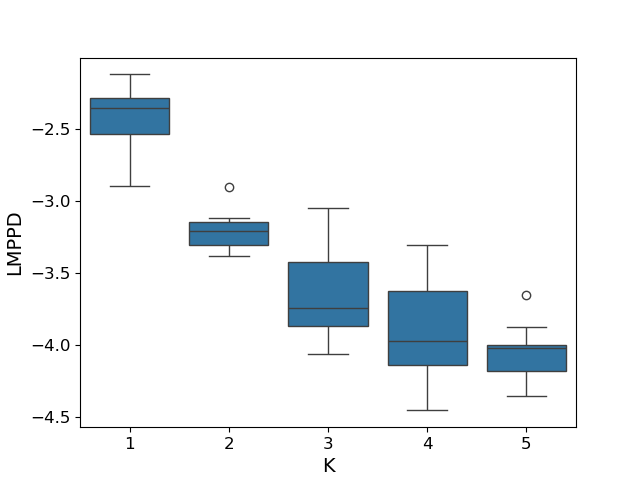}
        \caption{$\mc K = 1$}
    \end{subfigure}
    \hfill
    \begin{subfigure}[b]{0.23\textwidth}
        \centering
        \includegraphics[scale=0.26]{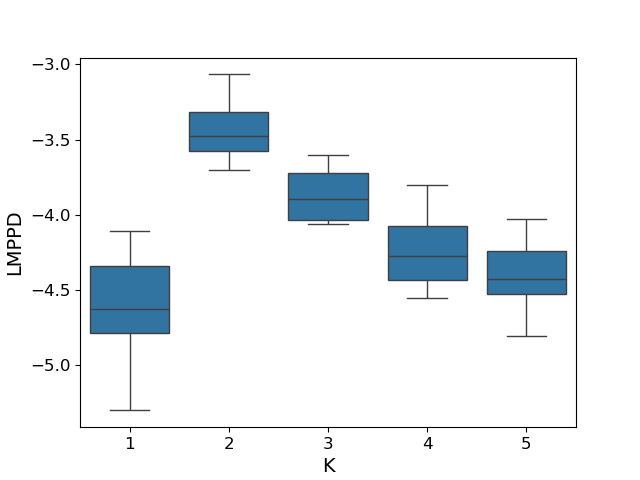}
        \caption{$\mc K = 2$}
    \end{subfigure}
    \hfill
    \begin{subfigure}[b]{0.23\textwidth}
        \centering
        \includegraphics[scale=0.26]{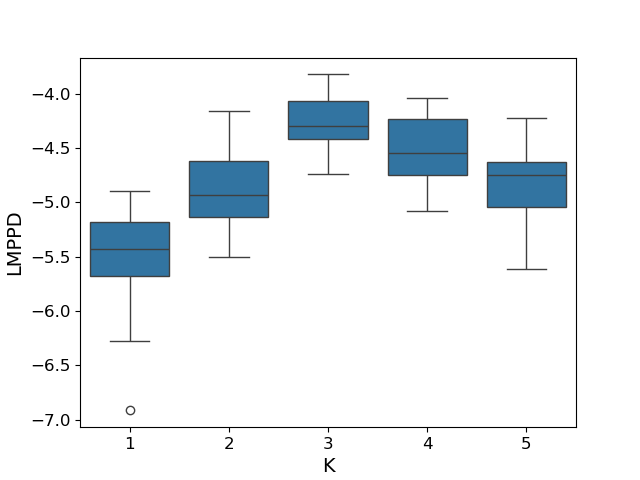}
        \caption{$\mc K = 3$}
    \end{subfigure}
    \hfill
    \begin{subfigure}[b]{0.23\textwidth}
        \centering
        \includegraphics[scale=0.26]{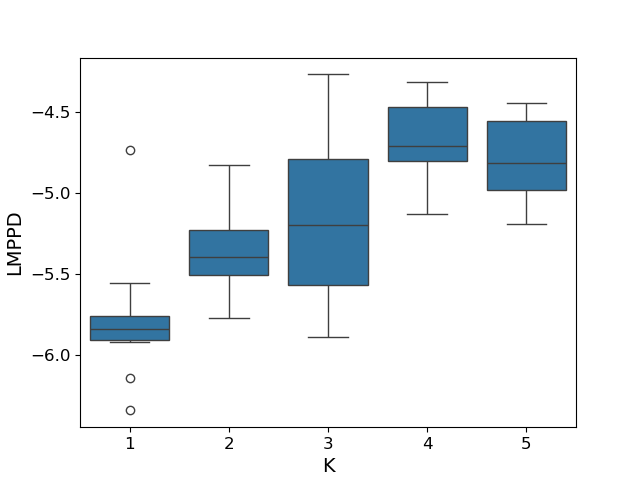}
        \caption{$\mc K = 4$}
    \end{subfigure}
    \caption{Boxplots of the log  marginal posterior predictive density LMPPD of 10 realizations from the model for $\mc K\in\{1,\ldots,4\}$, $\mc N=500$ $S=\mbox{Low}$, estimated using $K\in\{1,\ldots,5\}$.}
    \label{fig:cv_error}
\end{figure}

The results are presented in Figure~\ref{fig:cv_error}, where the four sub-figures correspond to the scenarios where the true model contained $\mc K=1\ldots4$ mixture components, respectively. Figure~\ref{fig:cv_error} shows that the proposed selection technique identifies the correct number of mixture components. 

Further to the above synthetic data experiments, we have compared the proposed methodology against naive baseline estimation methods, including: sequentially learning the clusters and then estimating each of the cluster graphs and removing the distinction between modifiable and non-modifiable covariates, thus incorporating all variables in the graph. Due to space constraints, we have incorporated this results in the Supplementary Material.

\begin{figure*}[tb]
    \centering
    \begin{tabular}{cc}
        \begin{subfigure}{0.3\linewidth}
            \centering
            \includegraphics[width=4cm]{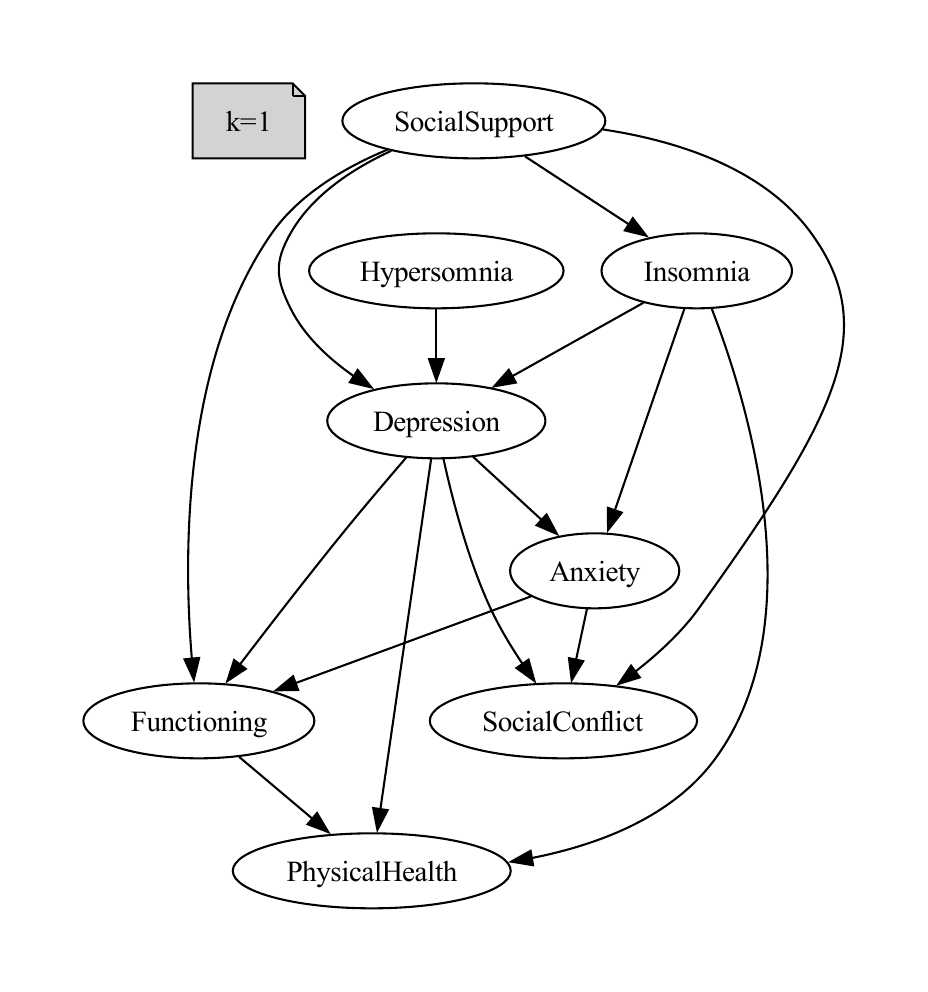}
        \end{subfigure} & \vline
        \begin{subfigure}{0.65\linewidth}
            \centering
            \includegraphics[width=6cm]{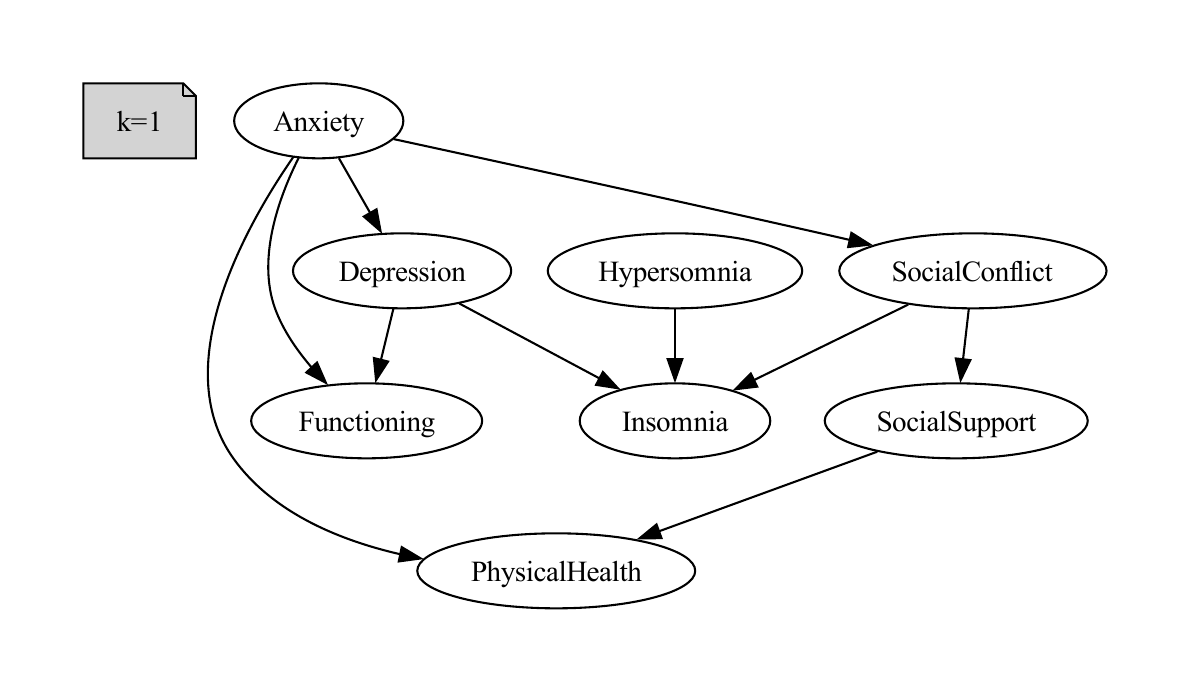}
            \includegraphics[width=4cm]{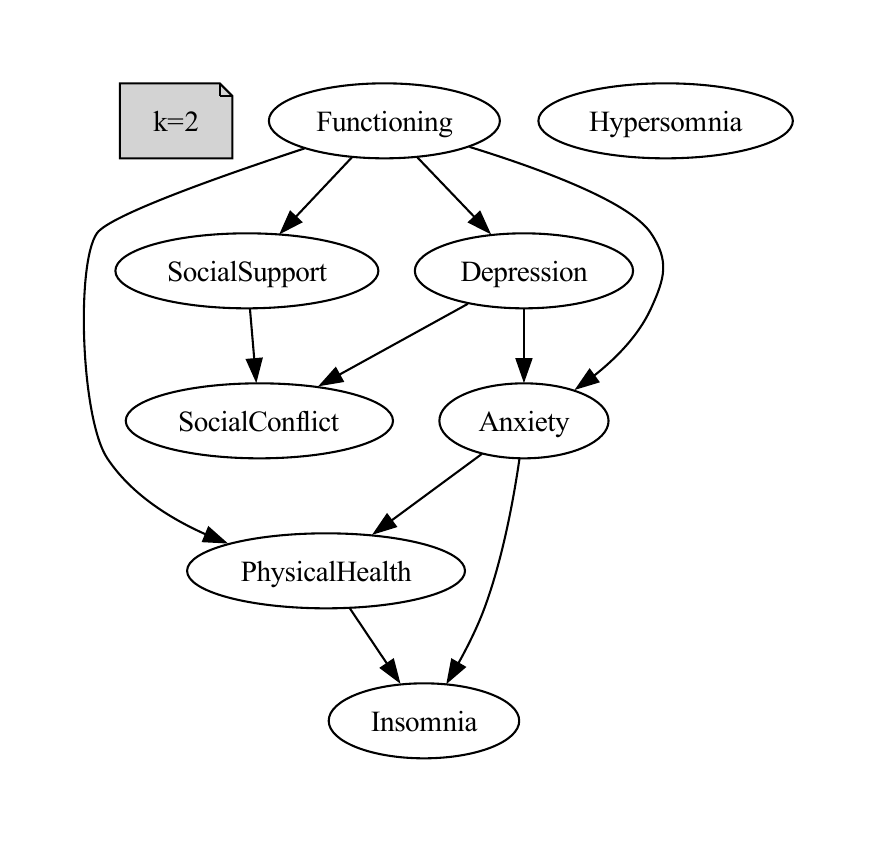}
        \end{subfigure} \\
        \hline
        \multicolumn{2}{c}{
            \begin{subfigure}{\linewidth}
                \centering
                \includegraphics[width=3.5cm]{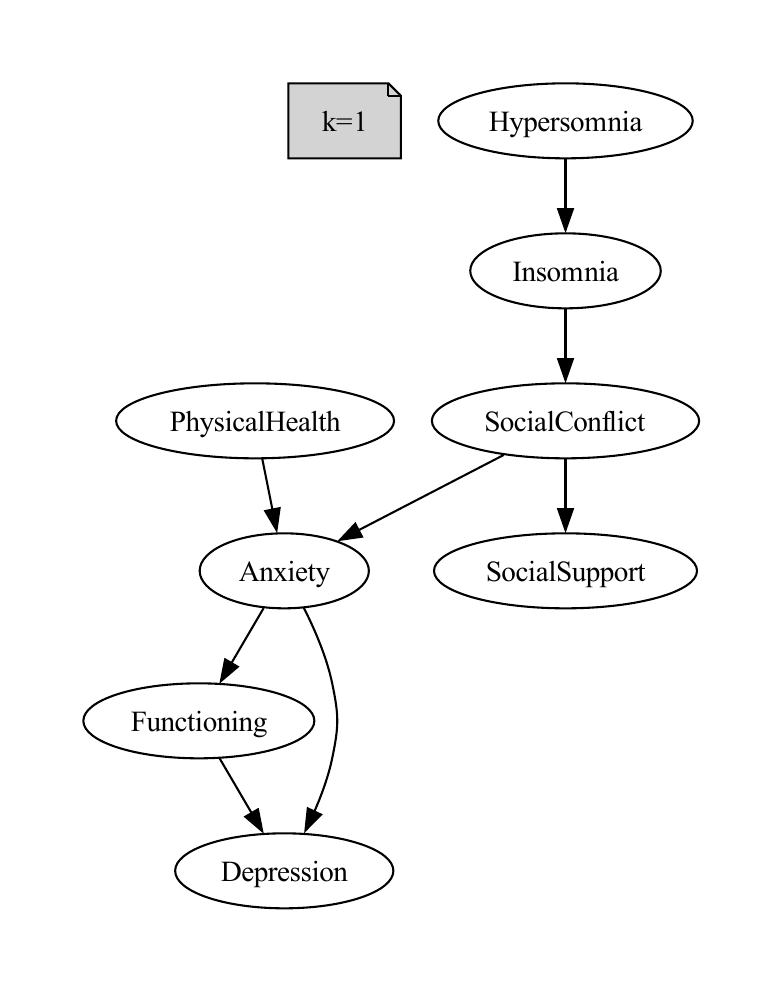}
                \includegraphics[width=5cm]{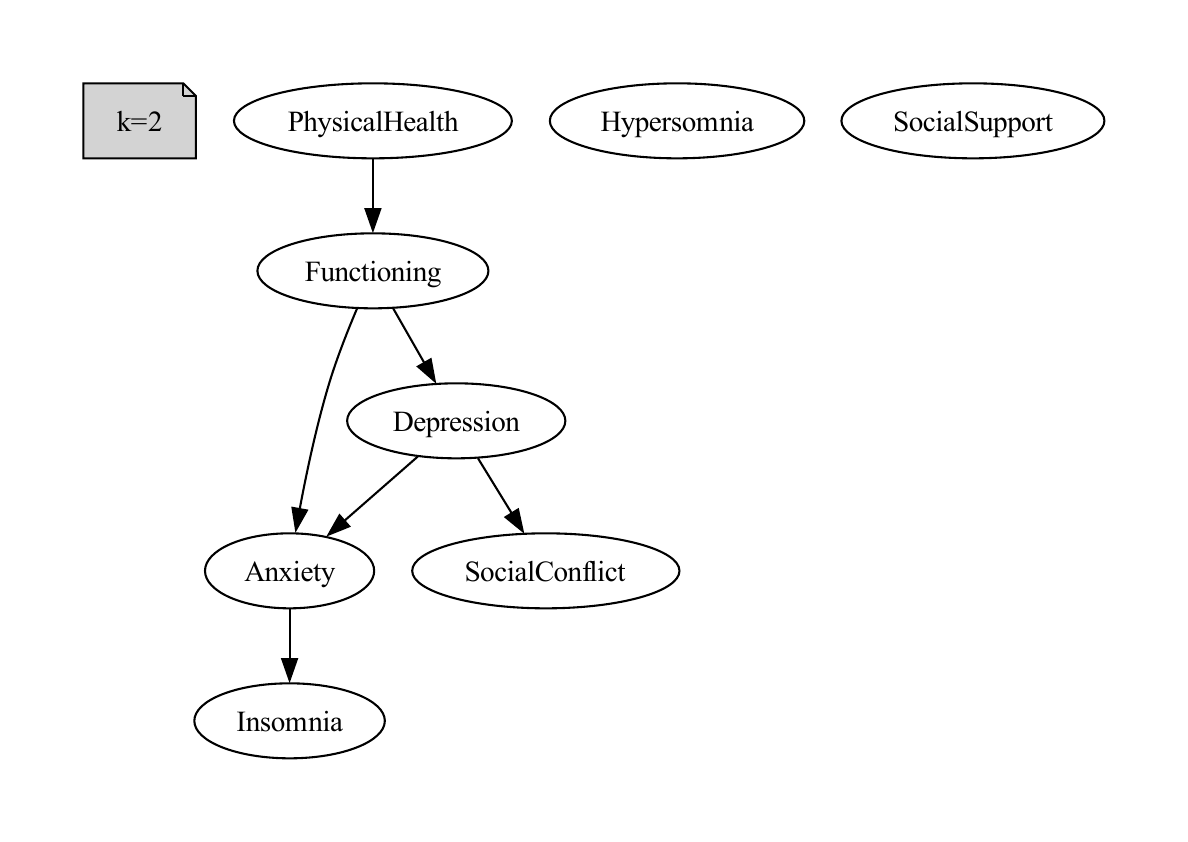}
                \includegraphics[width=5cm]{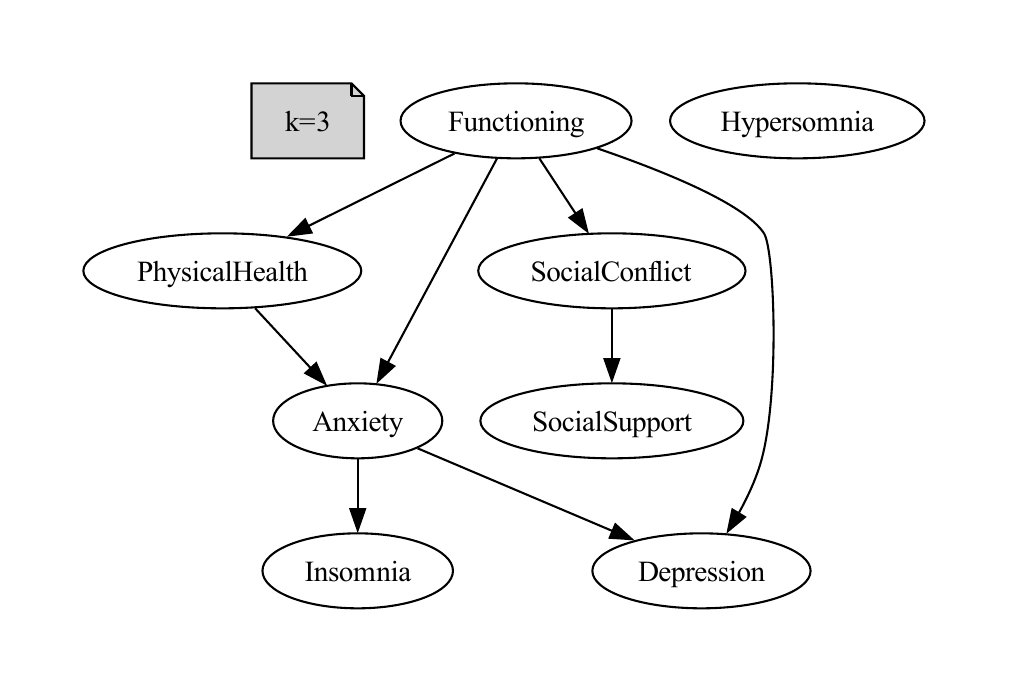}
            \end{subfigure}
        }\\
        \hline
        \multicolumn{2}{c}{
            \begin{subfigure}{\linewidth}
                \centering
                \includegraphics[width=4cm]{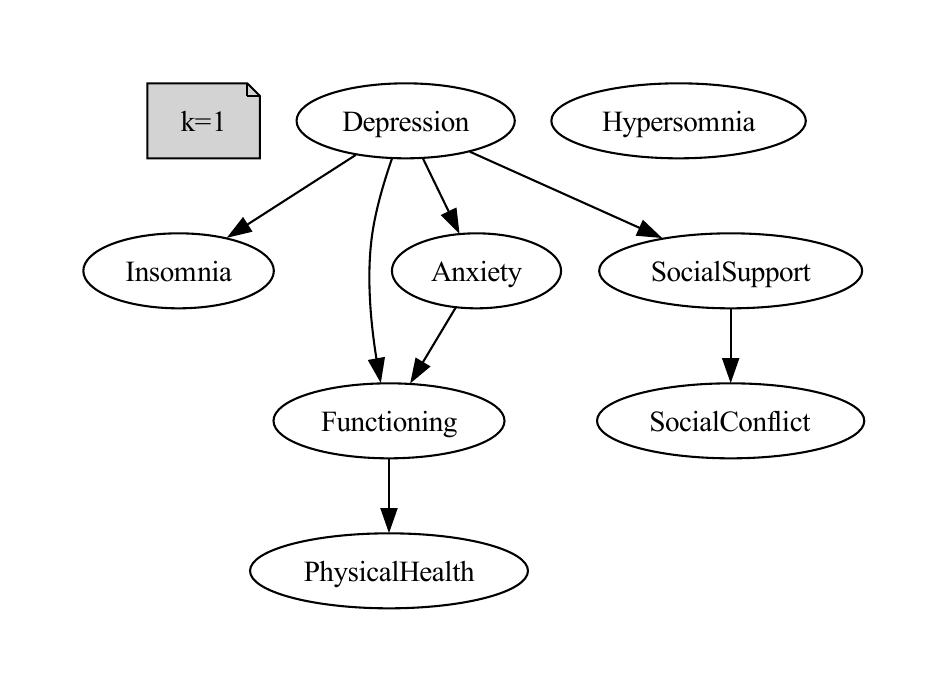}
                \includegraphics[width=4cm]{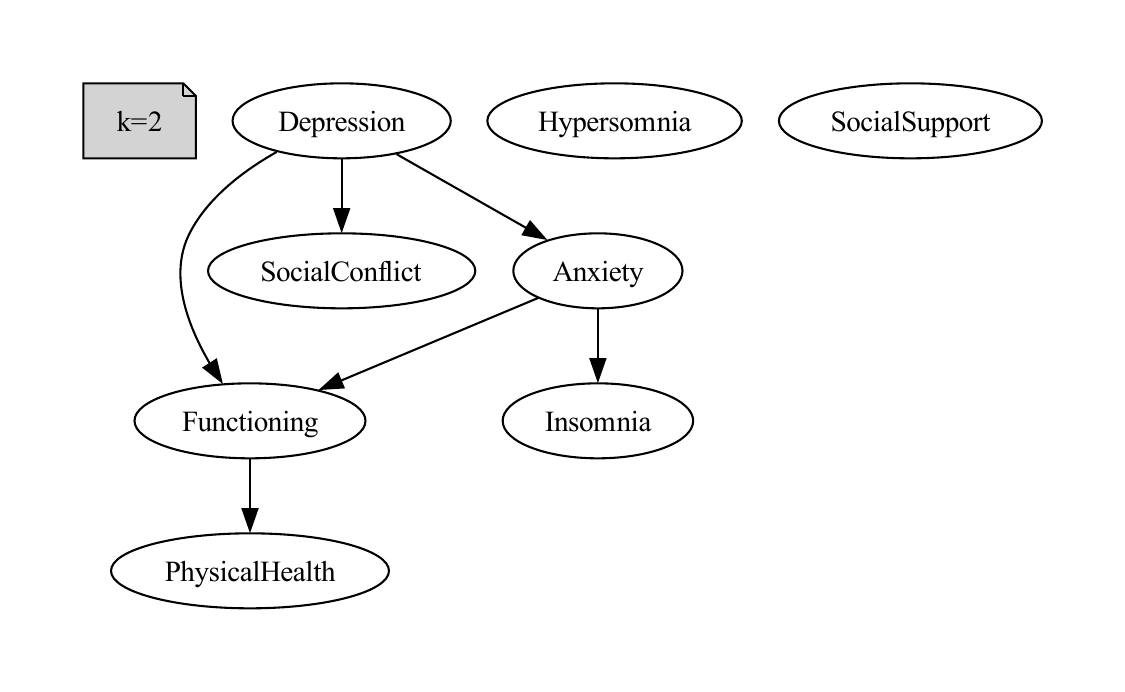}
                \includegraphics[width=4cm]{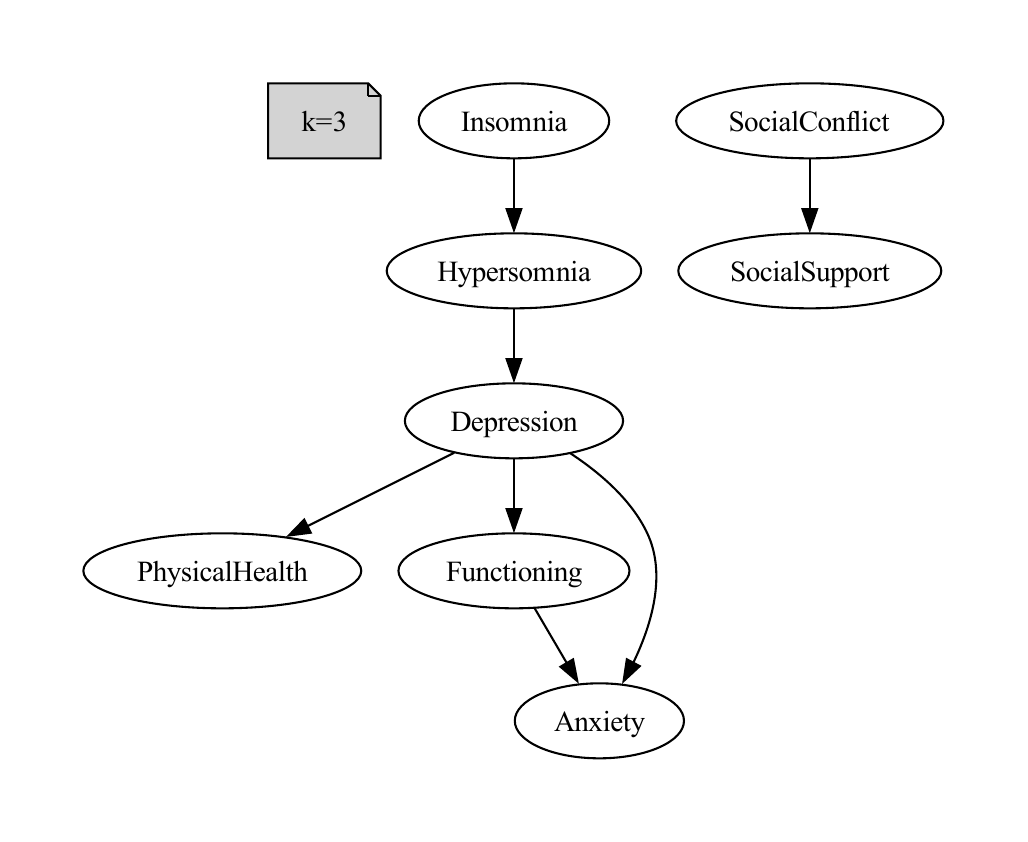}\\
                \includegraphics[width=7cm]{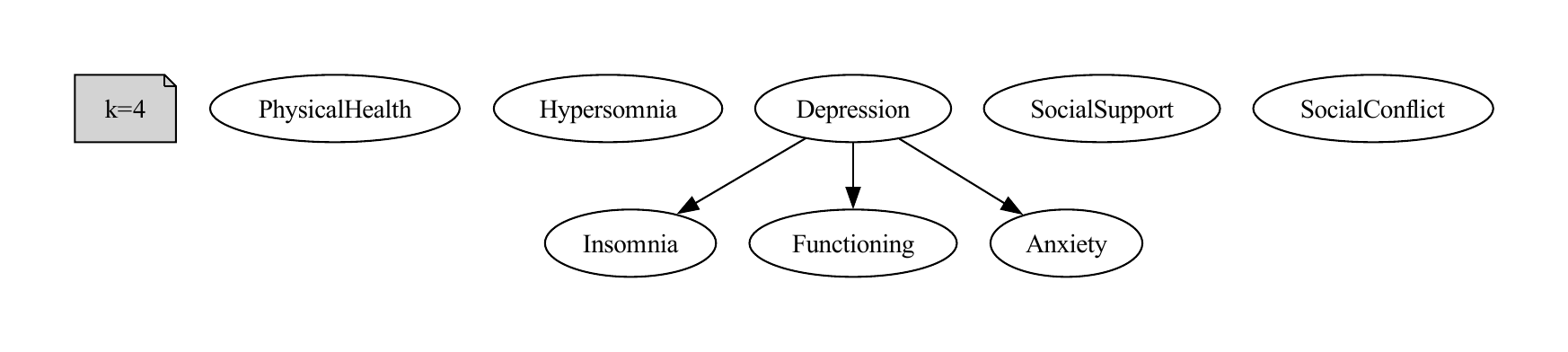}
            \end{subfigure}
        }
    \end{tabular}
    \caption{MAP Graphs for each cluster and total number of clusters}
    \label{fig:graphs}
\end{figure*}

\subsection{A Mental Health Case Study and Discussion}\label{sec:mental_health}

This experiment demonstrates the advantages of applying a flexible model to real-world data, particularly in suggesting a more informative causal landscape through its posterior inferences across graph structures, parameters, and mixture memberships. 

Mental illness affects 1 in 4 youth by 25 years of age and is the leading cause of disability and death \citep{jones_adult_2013, solmi_age_2022, colman_forty-year_2007}. Clinical services are designed to respond to this need, though the needs of young people are complex and span multiple domains (e.g.., social, physical, emotional, cognitive, environmental) \citep{patton_our_2016, kilbourne_framework_2010}. This makes individual treatment decisions difficult and leads to delayed, reactive, or inappropriate care. In this context understanding causal pathways within a heterogeneous population is crucial because it enables more precise decisions regarding treatment. 

Our dataset comprises 1565 individuals during their engagement with mental health services, with measures on eight factors deemed modifiable, and  four factors which are non-modifiable.  The modifiable features are Social Functioning (SF), Depression (Dep), Anxiety (Anx), Physical Health (PH), Social Support (SS), Social Conflict (SC), Insomnia (In), and Hypersomnia (Hyp). The non-modifiable features include Gender (G), Historic Traumatic Effect (TE), Historic Mental Health Issues (MHI), and Historic Suicidal Thoughts (ST). This dataset was split into 90\% train and 10\% test, allowing for evaluation of the LMPPD for out of sample data. We conducted experiments with mixture models comprising $K=\{1,2,3,4,5\}$, components. We ran Algorithm \ref{alg:mcmc} for 20,000 iterations, consuming a total of $5 [kSU]$. This computational effort was dedicated to evaluating the whole posterior distributions for different values of K.

\begin{table*}[tb]
    \centering
    \small 
    \caption{Posterior mean estimates of the mixing function weights, $\hat{E(\bs\beta)}_j\times 10^{-2}$ for $k=0,1, \ldots, K-1$. $\bs\beta \times 10^{-2}$. Note that K=1 case is essentially PMCMC over the entire dataset and therefore there are no mixing weights.}
    \begin{tabular}{|c|c|cc|ccc|}
        \hline
        \multirow{2}{*}{Non-modifiable} & \multicolumn{1}{c|}{$K=2$} & \multicolumn{2}{c|}{$K=3$} & \multicolumn{3}{c|}{$K=4$} \\
        \cline{2-7}
                          & $k=1$ & $k=1$ & $k=2$ & $k=1$ & $k=2$ & $k=3$\\
        \hline
        Intercept         & -0.42 & -4.9 & -4.2 & -6.8 & -6.0 & -6.8\\
        Gender            & -0.44 & -3.5 & -3.0 & -4.6 & -4.2 & -4.8\\
        Traumatic Event   & 0.73  & -1.3  & -2.7 & -3.4 & -2.6 & -3.2\\
        PrevMentalHealth  & 0.84  & -2.3  & -3.7 & -4.9 & -4.4 & -4.4\\
        SuicidalThoughts  & 1.24  & -1.4  & -3.7 & -4.4 & -3.4 & -4.1\\
        \hline
    \end{tabular}
    \label{tab:clusters}
\end{table*}

\begin{table*}[tb]
    \centering
    \small 
    \caption{Proportion of the population by non-modifiable features, $\mb{x}$, for each mixture component with index $k$, across models with total number of mixture components $K$}
    \begin{tabular}{|c|c|cc|ccc|cccc|}
        \hline
        \multirow{2}{*}{$\mathbf{x}$} & \multicolumn{1}{c|}{$K=1$} & \multicolumn{2}{c|}{$K=2$} & \multicolumn{3}{c|}{$K=3$} & \multicolumn{4}{c|}{$K=4$} \\
        \cline{2-11}
        & $k=1$ & $k=1$ & $k=2$ & $k=1$ & $k=2$ & $k=3$ & $k=1$ & $k=2$ & $k=3$ & $k=4$ \\
        \hline
        N     & 1408  & 712 & 696  & 500 & 476 & 431 & 428 & 350 & 357 & 273\\
        \hline
        G     & 0.71  & 0.72 & 0.70  & 0.74 & 0.68 & 0.71 & 0.71 & 0.69 & 0.77 & 0.66 \\
        TE    & 0.49  & 0.57 & 0.40  & 0.61 & 0.43 & 0.40 & 0.40 & 0.43 & 0.68 & 0.44 \\
        MHI   & 0.70  & 0.78 & 0.62  & 0.80 & 0.65 & 0.65 & 0.64 & 0.63 & 0.81 & 0.75 \\
        ST    & 0.60  & 0.75 & 0.45  & 0.80 & 0.51 & 0.47 & 0.47 & 0.47 & 0.80 & 0.72 \\
        \hline
    \end{tabular}
    \label{tab:summary_statistics}
\end{table*}

\begin{figure}[tb]
    \centering
        \centering
        \includegraphics[width=0.85\linewidth]{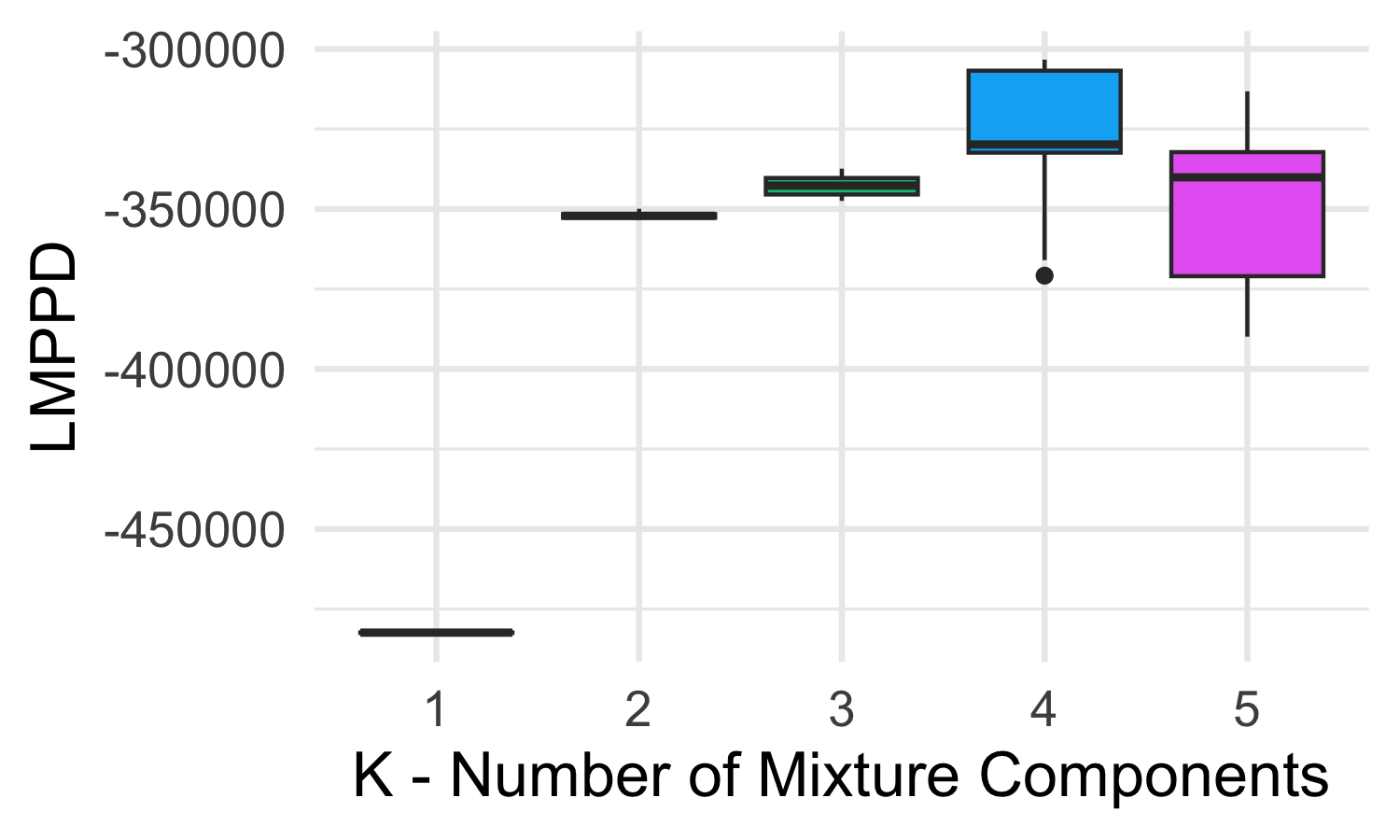}
    \caption{Total log marginal predictive probability density (LMPPD) for the test data over a varying number of mixture components $K$. Each box summarises the scores over 10 different runs of the experiments with different starting DAGs.}
    \label{fig:log_prob_test_mh}
\end{figure}

\begin{figure}[tb]
    \centering
        \centering
        \includegraphics[width=0.85\linewidth]{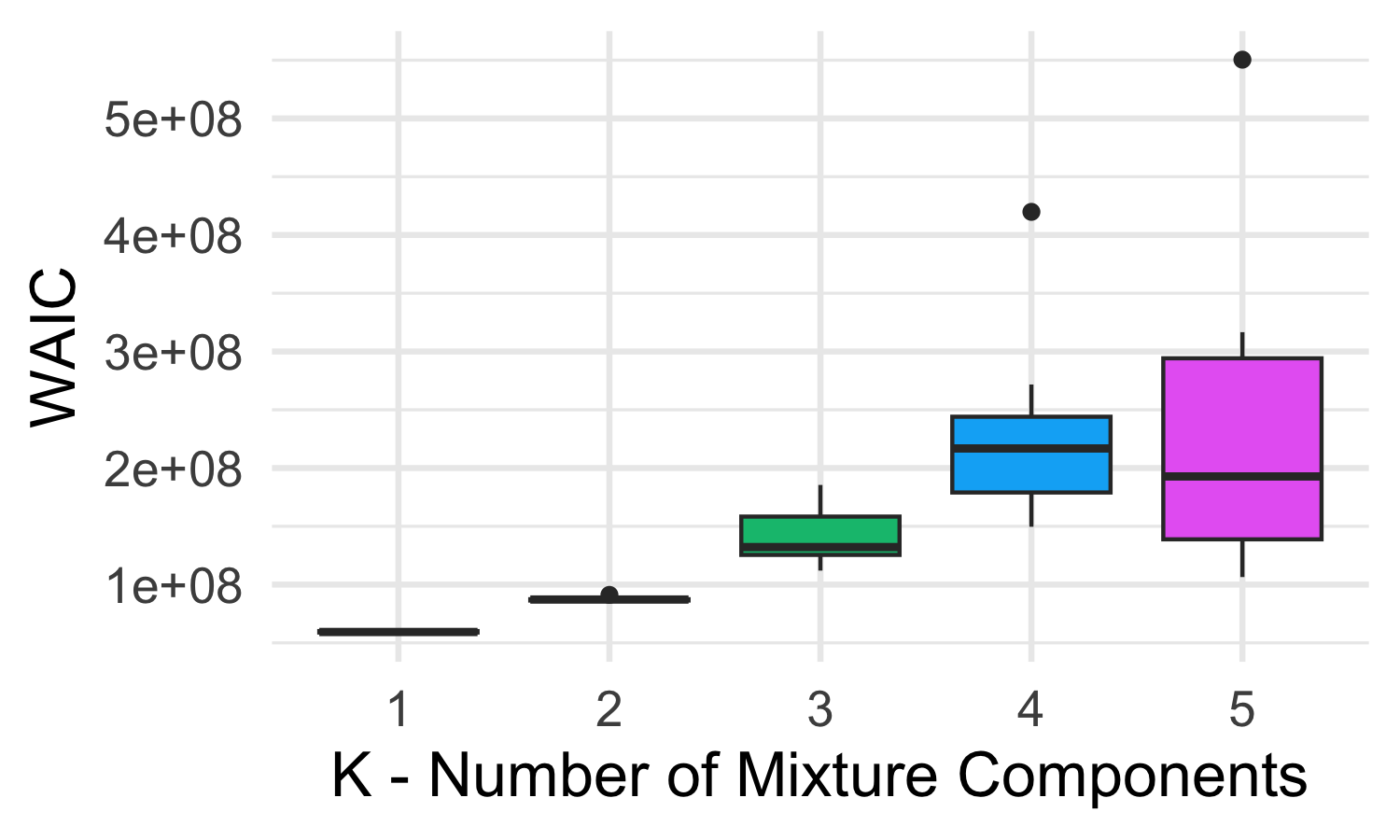}
    \caption{WAIC scores over in-sample data for the real-world case study in \ref{sec:mental_health}. For each mixture component $K$, 10 MCMC runs were conducted, each initialized with a different starting DAG.}
    \label{fig:waic_innowell}
\end{figure}

 The {\it Maximum-A-Posteriori} (MAP) graphs presented in Figure~\ref{fig:graphs}, together with the LMPPD in Figure~\ref{fig:log_prob_test_mh} demonstrate how this mixtures methodology can provide useful insight into the distinct underlying processes that give rise to specific mental health presentations which could inform specific treatment decisions. Figure~\ref{fig:log_prob_test_mh} shows a large improvement in LMPPD in going from $K=1$ to $K=2$, demonstrating strong evidence for heterogeneity in the population. The recognised heterogeneity of depression is a central reason why so many studies of causation or biomarkers have failed to yield significant replicable findings that are clinically useful \citep{hickie_what_2024}. This has led to major calls for improved stratification of mental disorders with an emphasis on new methodologies to identify relevant subgroups based on causal processes which could be used to guide specific interventions. 
 
 Although the highest LMPPD corresponds to $K=4$ the improvements in going from $K=2$, and $K=3$ and from $K=3$ to $K=4$ are less pronounced, and the results from all three settings, $K=2,3,4$, warrant discussion. For $K=5$ the LMPPD decreased, which indicated overfitting.

 Additionally, the {\it Watanabe–Akaike Information Criterion} (WAIC) \citep{watanabe2013widely} was calculated to evaluate the predictive performance of the models, while accounting for larger complexity as $K$ increases. The WAIC for varying $K$ are summarized in Figure~\ref{fig:waic_innowell}. For each $K$, 10 different MCMC chains were computed with randomized initial DAGs. The decrease in the WAIC score for $K>4$ agrees with the LMPPD results presented in Figure~\ref{fig:log_prob_test_mh}. Confirming that a selection of 4 mixture components is optimal for this specific dataset, considering both in-sample (WAIC) and out-of-sample (LMPPD) evaluations.

Panels (b), (c) and (d) of Figure \ref{fig:graphs} show the emergence of  distinct, probable causal processes that would lead to very different treatment decisions. In panel (b), we see the critical role of anxiety in the development of depression and insomnia, as well as its impacts on social and occupational function. This is consistent with a hypothesised anxious-depression pathway in youth mental health whereby elevated activity to the sympathetic nervous system and hypothalamus-pituitary-adrenal axis is a critical causal mechanism driving mantel health and other poor health outcomes \citep{hickie_right_2019,iorfino_social_2022}. By contrast, in panel (b, k=2), we see a distinct set of processes, largely driven by functioning and the inverse relationship between depression and anxiety which would be consistent with other mechanisms linked to dysregulated mood, cognition and activation. The intervention decisions for each of these proposed subgroups would be vastly different as they link to distinct pathophysiological mechanisms that would have differential treatment response effects. Importantly, interventions that may be useful for one group (e.g., antidepressant medication), may be harmful for the other group due to distinct underlying processes. This work demonstrates the utility of this approach to identify more homogenous subgroups that are comparable in the underlying process driving illness. This type of stratification approach is needed in mental health to drive more personalised intervention decisions that are safe and relevant to a person’s specific needs.

Tables~\ref{tab:summary_statistics} shows the summary statistics for the non-modifiable variables and indicates that the strongest predictor of cluster membership is previous suicidal thoughts (ST), followed previous mental health issue (MHI) and prior traumatic event (TE).  Gender does not seem to play a role at all.  The proportion of the total population, $K=1$ with previous suicidal thoughts is 0.60, this quickly segreggates into 0.75 and 0.45 for  the two components in $K=2$. A similiar pattern is evident for MHI and TE.  Interestingly it is TE which appears to be driving the likelihood of a $K=4$ component mixture, with one of the four clusters, $k=3$, having a much higher proportion of individuas with previous traumatic event than the other three clusters.

\section{Conclusion}\label{sec:conclusion}
In this paper we have presented a technique for estimating and inferring multiple graph structures which result from heterogeneous data generating processes (DGPs) that exist within a population. We have shown, via simulation, that our method can estimate the most likely number of DGPs within a population and that it can recover the most likely graphs structures for those DGPs. Most important, the method shows how this is applicable to a real example for youth mental health and how the insights from the proposed method may guide treatment. While the methodology offers significant benefits, careful consideration of potential risks, such as privacy concerns, bias, and misuse, is essential to ensure ethical and responsible application.

We note that the division of modifiable and non-modifiable factors is subjective, which may be considered a limitation. However, we consider it a useful way in which to incorporate domain knowledge into the process. 

Other limitations, which will be the subject of future work, include extending the method to allow for the number of possible clusters to be random variable, so that inference for, rather than selection of, this parameter is available.

\bibliography{bibliography}


\begin{thebibliography}{33}


\ifx \showCODEN    \undefined \def \showCODEN     #1{\unskip}     \fi
\ifx \showDOI      \undefined \def \showDOI       #1{#1}\fi
\ifx \showISBNx    \undefined \def \showISBNx     #1{\unskip}     \fi
\ifx \showISBNxiii \undefined \def \showISBNxiii  #1{\unskip}     \fi
\ifx \showISSN     \undefined \def \showISSN      #1{\unskip}     \fi
\ifx \showLCCN     \undefined \def \showLCCN      #1{\unskip}     \fi
\ifx \shownote     \undefined \def \shownote      #1{#1}          \fi
\ifx \showarticletitle \undefined \def \showarticletitle #1{#1}   \fi
\ifx \showURL      \undefined \def \showURL       {\relax}        \fi
\providecommand\bibfield[2]{#2}
\providecommand\bibinfo[2]{#2}
\providecommand\natexlab[1]{#1}
\providecommand\showeprint[2][]{arXiv:#2}

\bibitem[Anderson(2009)]%
        {anderson2009}
\bibfield{author}{\bibinfo{person}{T.W. Anderson}.}
  \bibinfo{year}{2009}\natexlab{}.
\newblock \bibinfo{booktitle}{\emph{An Introduction to Multivariate Statistical
  Analysis, 3rd Ed}}.
\newblock \bibinfo{publisher}{Wiley India Pvt. Limited}.
\newblock
\showISBNx{9788126524488}
\urldef\tempurl%
\url{https://books.google.com.au/books?id=1iF0CgAAQBAJ}
\showURL{%
\tempurl}


\bibitem[Castelletti et~al\mbox{.}(2020)]%
        {Castelletti2020}
\bibfield{author}{\bibinfo{person}{Federico Castelletti}, \bibinfo{person}{Luca
  La~Rocca}, \bibinfo{person}{Stefano Peluso}, \bibinfo{person}{Francesco~C.
  Stingo}, {and} \bibinfo{person}{Guido Consonni}.}
  \bibinfo{year}{2020}\natexlab{}.
\newblock \showarticletitle{Bayesian learning of multiple directed networks
  from observational data}.
\newblock \bibinfo{journal}{\emph{Statistics in Medicine}}
  \bibinfo{volume}{39}, \bibinfo{number}{30} (\bibinfo{year}{2020}),
  \bibinfo{pages}{4745--4766}.
\newblock
\urldef\tempurl%
\url{https://doi.org/10.1002/sim.8751}
\showDOI{\tempurl}
\showeprint{https://onlinelibrary.wiley.com/doi/pdf/10.1002/sim.8751}


\bibitem[Chen et~al\mbox{.}(2022)]%
        {chen2022}
\bibfield{author}{\bibinfo{person}{Zixiang Chen}, \bibinfo{person}{Yihe Deng},
  \bibinfo{person}{Yue Wu}, \bibinfo{person}{Quanquan Gu}, {and}
  \bibinfo{person}{Yuanzhi Li}.} \bibinfo{year}{2022}\natexlab{}.
\newblock \showarticletitle{Towards understanding the mixture-of-experts layer
  in deep learning}.
\newblock \bibinfo{journal}{\emph{Advances in neural information processing
  systems}}  \bibinfo{volume}{35} (\bibinfo{year}{2022}),
  \bibinfo{pages}{23049--23062}.
\newblock


\bibitem[Chobtham and Constantinou(2020)]%
        {Chobtham2020}
\bibfield{author}{\bibinfo{person}{Kiattikun Chobtham} {and}
  \bibinfo{person}{Anthony~C. Constantinou}.} \bibinfo{year}{2020}\natexlab{}.
\newblock \showarticletitle{Bayesian network structure learning with causal
  effects in the presence of latent variables}. In
  \bibinfo{booktitle}{\emph{Proceedings of the 10th International Conference on
  Probabilistic Graphical Models}} \emph{(\bibinfo{series}{Proceedings of
  Machine Learning Research}, Vol.~\bibinfo{volume}{138})},
  \bibfield{editor}{\bibinfo{person}{Manfred Jaeger} {and}
  \bibinfo{person}{Thomas~Dyhre Nielsen}} (Eds.). \bibinfo{publisher}{PMLR},
  \bibinfo{pages}{101--112}.
\newblock
\urldef\tempurl%
\url{https://proceedings.mlr.press/v138/chobtham20a.html}
\showURL{%
\tempurl}


\bibitem[Chobtham and Constantinou(2022)]%
        {Chobtham2022}
\bibfield{author}{\bibinfo{person}{Kiattikun Chobtham} {and}
  \bibinfo{person}{Anthony~C. Constantinou}.} \bibinfo{year}{2022}\natexlab{}.
\newblock \showarticletitle{Discovery and density estimation of latent
  confounders in Bayesian networks with evidence lower bound}. In
  \bibinfo{booktitle}{\emph{Proceedings of The 11th International Conference on
  Probabilistic Graphical Models}} \emph{(\bibinfo{series}{Proceedings of
  Machine Learning Research}, Vol.~\bibinfo{volume}{186})},
  \bibfield{editor}{\bibinfo{person}{Antonio Salmeron} {and}
  \bibinfo{person}{Rafael Rumi}} (Eds.). \bibinfo{publisher}{PMLR},
  \bibinfo{pages}{121--132}.
\newblock
\urldef\tempurl%
\url{https://proceedings.mlr.press/v186/chobtham22a.html}
\showURL{%
\tempurl}


\bibitem[Colman et~al\mbox{.}(2007)]%
        {colman_forty-year_2007}
\bibfield{author}{\bibinfo{person}{Ian Colman}, \bibinfo{person}{Michael E~J
  Wadsworth}, \bibinfo{person}{Tim~J Croudace}, {and} \bibinfo{person}{Peter~B
  Jones}.} \bibinfo{year}{2007}\natexlab{}.
\newblock \showarticletitle{Forty-{Year} {Psychiatric} {Outcomes} {Following}
  {Assessment} for {Internalizing} {Disorder} in {Adolescence}}.
\newblock \bibinfo{journal}{\emph{Am J Psychiatry}} (\bibinfo{year}{2007}).
\newblock


\bibitem[Dawid(2010)]%
        {dawid2010}
\bibfield{author}{\bibinfo{person}{A.~Philip Dawid}.}
  \bibinfo{year}{2010}\natexlab{}.
\newblock \showarticletitle{Beware of the DAG!}. In
  \bibinfo{booktitle}{\emph{Proceedings of Workshop on Causality: Objectives
  and Assessment at NIPS 2008}} \emph{(\bibinfo{series}{Proceedings of Machine
  Learning Research}, Vol.~\bibinfo{volume}{6})},
  \bibfield{editor}{\bibinfo{person}{Isabelle Guyon}, \bibinfo{person}{Dominik
  Janzing}, {and} \bibinfo{person}{Bernhard Schölkopf}} (Eds.).
  \bibinfo{publisher}{PMLR}, \bibinfo{address}{Whistler, Canada},
  \bibinfo{pages}{59--86}.
\newblock


\bibitem[Dawid(2024)]%
        {dawid2024}
\bibfield{author}{\bibinfo{person}{Philip Dawid}.}
  \bibinfo{year}{2024}\natexlab{}.
\newblock \bibinfo{title}{What Is a Causal Graph?}
\newblock
\newblock
\showeprint[arxiv]{2402.09429}~[math.ST]


\bibitem[Erd\"os and R\'enyi(1959)]%
        {Erdos1959}
\bibfield{author}{\bibinfo{person}{P Erd\"os} {and} \bibinfo{person}{A
  R\'enyi}.} \bibinfo{year}{1959}\natexlab{}.
\newblock \showarticletitle{On Random Graphs I}.
\newblock \bibinfo{journal}{\emph{Publicationes Mathematicae Debrecen}}
  \bibinfo{volume}{6} (\bibinfo{year}{1959}), \bibinfo{pages}{290--297}.
\newblock


\bibitem[Geiger and Heckerman(2002)]%
        {Geiger2002}
\bibfield{author}{\bibinfo{person}{Dan Geiger} {and} \bibinfo{person}{David
  Heckerman}.} \bibinfo{year}{2002}\natexlab{}.
\newblock \showarticletitle{Parameter Priors for Directed Acyclic Graphical
  Models and the Characterization of Several Probability Distributions}.
\newblock \bibinfo{journal}{\emph{The Annals of Statistics}}
  \bibinfo{volume}{30}, \bibinfo{number}{5} (\bibinfo{year}{2002}),
  \bibinfo{pages}{1412--1440}.
\newblock
\showISSN{00905364}
\urldef\tempurl%
\url{http://www.jstor.org/stable/1558719}
\showURL{%
\tempurl}


\bibitem[Geiger and Heckerman(2021)]%
        {geiger2021}
\bibfield{author}{\bibinfo{person}{Dan Geiger} {and} \bibinfo{person}{David
  Heckerman}.} \bibinfo{year}{2021}\natexlab{}.
\newblock \bibinfo{title}{Parameter Priors for Directed Acyclic Graphical
  Models and the Characterization of Several Probability Distributions}.
\newblock
\newblock
\showeprint[arxiv]{2105.03248}~[stat.ML]


\bibitem[Green(1995)]%
        {green95}
\bibfield{author}{\bibinfo{person}{PETER~J. Green}.}
  \bibinfo{year}{1995}\natexlab{}.
\newblock \showarticletitle{{Reversible jump Markov chain Monte Carlo
  computation and Bayesian model determination}}.
\newblock \bibinfo{journal}{\emph{Biometrika}} \bibinfo{volume}{82},
  \bibinfo{number}{4} (\bibinfo{date}{12} \bibinfo{year}{1995}),
  \bibinfo{pages}{711--732}.
\newblock
\showISSN{0006-3444}
\urldef\tempurl%
\url{https://doi.org/10.1093/biomet/82.4.711}
\showDOI{\tempurl}
\showeprint{https://academic.oup.com/biomet/article-pdf/82/4/711/699533/82-4-711.pdf}


\bibitem[H{\"a}ggstr{\"o}m(2018)]%
        {haggstrom2018}
\bibfield{author}{\bibinfo{person}{Jenny H{\"a}ggstr{\"o}m}.}
  \bibinfo{year}{2018}\natexlab{}.
\newblock \showarticletitle{Data-driven confounder selection via Markov and
  Bayesian networks}.
\newblock \bibinfo{journal}{\emph{Biometrics}} \bibinfo{volume}{74},
  \bibinfo{number}{2} (\bibinfo{year}{2018}), \bibinfo{pages}{389--398}.
\newblock


\bibitem[Hickie et~al\mbox{.}(2024)]%
        {hickie_what_2024}
\bibfield{author}{\bibinfo{person}{Ian~B. Hickie}, \bibinfo{person}{Michael
  Berk}, \bibinfo{person}{Jan Scott}, \bibinfo{person}{Jacob Crouse},
  \bibinfo{person}{Elizabeth Scott}, \bibinfo{person}{Naomi Wray}, {and}
  \bibinfo{person}{Frank Iorfino}.} \bibinfo{year}{2024}\natexlab{}.
\newblock \showarticletitle{What are the best strategies for stratification of
  clinical cohorts with depression and other mood disorders?}
\newblock \bibinfo{journal}{\emph{Research Directions: Depression}}
  \bibinfo{volume}{1} (\bibinfo{year}{2024}), \bibinfo{pages}{e18}.
\newblock
\showISSN{2976-9000}
\urldef\tempurl%
\url{https://doi.org/10.1017/dep.2024.1}
\showDOI{\tempurl}


\bibitem[Hickie et~al\mbox{.}(2019)]%
        {hickie_right_2019}
\bibfield{author}{\bibinfo{person}{Ian~B Hickie}, \bibinfo{person}{Elizabeth~M
  Scott}, \bibinfo{person}{Shane~P Cross}, \bibinfo{person}{Frank Iorfino},
  \bibinfo{person}{Tracey~A Davenport}, \bibinfo{person}{Adam~J Guastella},
  \bibinfo{person}{Sharon~L Naismith}, \bibinfo{person}{Joanne~S Carpenter},
  \bibinfo{person}{Cathrin Rohleder}, \bibinfo{person}{Jacob~J Crouse},
  \bibinfo{person}{Daniel~F Hermens}, \bibinfo{person}{Dagmar Koethe},
  \bibinfo{person}{F Markus~Leweke}, \bibinfo{person}{Ashleigh~M Tickell},
  \bibinfo{person}{Vilas Sawrikar}, {and} \bibinfo{person}{Jan Scott}.}
  \bibinfo{year}{2019}\natexlab{}.
\newblock \showarticletitle{Right care, first time: a highly personalised and
  measurement‐based care model to manage youth mental health}.
\newblock \bibinfo{journal}{\emph{Medical Journal of Australia}}
  \bibinfo{volume}{211}, \bibinfo{number}{S9} (\bibinfo{date}{Nov.}
  \bibinfo{year}{2019}).
\newblock
\showISSN{0025-729X, 1326-5377}
\urldef\tempurl%
\url{https://doi.org/10.5694/mja2.50383}
\showDOI{\tempurl}


\bibitem[Higashigaki et~al\mbox{.}(2010)]%
        {Higashigaki2010}
\bibfield{author}{\bibinfo{person}{Tomoya Higashigaki}, \bibinfo{person}{Kaname
  Kojima}, \bibinfo{person}{Rui Yamaguchi}, \bibinfo{person}{Masato Inoue},
  \bibinfo{person}{Seiya Imoto}, {and} \bibinfo{person}{Satoru Miyano}.}
  \bibinfo{year}{2010}\natexlab{}.
\newblock \showarticletitle{Identifying Hidden Confounders in Gene Networks by
  Bayesian Networks}. In \bibinfo{booktitle}{\emph{2010 IEEE International
  Conference on BioInformatics and BioEngineering}}. \bibinfo{pages}{168--173}.
\newblock
\urldef\tempurl%
\url{https://doi.org/10.1109/BIBE.2010.35}
\showDOI{\tempurl}


\bibitem[Iorfino et~al\mbox{.}(2022)]%
        {iorfino_social_2022}
\bibfield{author}{\bibinfo{person}{Frank Iorfino}, \bibinfo{person}{Elizabeth~M
  Scott}, {and} \bibinfo{person}{Ian~B Hickie}.}
  \bibinfo{year}{2022}\natexlab{}.
\newblock \showarticletitle{Social and occupational outcomes for young people
  who attend early intervention mental health services: a longitudinal study}.
\newblock \bibinfo{journal}{\emph{Medical Journal of Australia}}
  \bibinfo{volume}{217}, \bibinfo{number}{4} (\bibinfo{date}{Aug.}
  \bibinfo{year}{2022}), \bibinfo{pages}{218--218}.
\newblock
\showISSN{0025-729X, 1326-5377}
\urldef\tempurl%
\url{https://doi.org/10.5694/mja2.51653}
\showDOI{\tempurl}


\bibitem[Jacobs(1995)]%
        {ME1995}
\bibfield{author}{\bibinfo{person}{Robert~A. Jacobs}.}
  \bibinfo{year}{1995}\natexlab{}.
\newblock \showarticletitle{Methods For Combining Experts' Probability
  Assessments}.
\newblock \bibinfo{journal}{\emph{Neural Computation}} \bibinfo{volume}{7},
  \bibinfo{number}{5} (\bibinfo{year}{1995}), \bibinfo{pages}{867--888}.
\newblock
\urldef\tempurl%
\url{https://doi.org/10.1162/neco.1995.7.5.867}
\showDOI{\tempurl}


\bibitem[Jones(2013)]%
        {jones_adult_2013}
\bibfield{author}{\bibinfo{person}{P.~B. Jones}.}
  \bibinfo{year}{2013}\natexlab{}.
\newblock \showarticletitle{Adult mental health disorders and their age at
  onset}.
\newblock \bibinfo{journal}{\emph{British Journal of Psychiatry}}
  \bibinfo{volume}{202}, \bibinfo{number}{s54} (\bibinfo{date}{Jan.}
  \bibinfo{year}{2013}), \bibinfo{pages}{s5--s10}.
\newblock
\showISSN{0007-1250, 1472-1465}
\urldef\tempurl%
\url{https://doi.org/10.1192/bjp.bp.112.119164}
\showDOI{\tempurl}


\bibitem[Kilbourne et~al\mbox{.}(2010)]%
        {kilbourne_framework_2010}
\bibfield{author}{\bibinfo{person}{A.~M. Kilbourne}, \bibinfo{person}{C.
  Fullerton}, \bibinfo{person}{D. Dausey}, \bibinfo{person}{H.~A. Pincus},
  {and} \bibinfo{person}{R.~C. Hermann}.} \bibinfo{year}{2010}\natexlab{}.
\newblock \showarticletitle{A framework for measuring quality and promoting
  accountability across silos: the case of mental disorders and co-occurring
  conditions}.
\newblock \bibinfo{journal}{\emph{Quality and Safety in Health Care}}
  \bibinfo{volume}{19}, \bibinfo{number}{2} (\bibinfo{date}{April}
  \bibinfo{year}{2010}), \bibinfo{pages}{113--116}.
\newblock
\showISSN{1475-3898, 1475-3901}
\urldef\tempurl%
\url{https://doi.org/10.1136/qshc.2008.027706}
\showDOI{\tempurl}


\bibitem[Koller and Friedman(2009)]%
        {koller2009}
\bibfield{author}{\bibinfo{person}{D. Koller} {and} \bibinfo{person}{N.
  Friedman}.} \bibinfo{year}{2009}\natexlab{}.
\newblock \bibinfo{booktitle}{\emph{Probabilistic Graphical Models: Principles
  and Techniques}}.
\newblock \bibinfo{publisher}{MIT Press}.
\newblock
\showISBNx{9780262013192}


\bibitem[Kuipers and Moffa(2017)]%
        {kuipers2017}
\bibfield{author}{\bibinfo{person}{Jack Kuipers} {and} \bibinfo{person}{Giusi
  Moffa}.} \bibinfo{year}{2017}\natexlab{}.
\newblock \showarticletitle{Partition {MCMC} for Inference on Acyclic
  Digraphs}.
\newblock \bibinfo{journal}{\emph{J. Amer. Statist. Assoc.}}
  \bibinfo{volume}{112}, \bibinfo{number}{517} (\bibinfo{date}{jan}
  \bibinfo{year}{2017}), \bibinfo{pages}{282--299}.
\newblock
\urldef\tempurl%
\url{https://doi.org/10.1080/01621459.2015.1133426}
\showDOI{\tempurl}


\bibitem[Oldfield et~al\mbox{.}(2024)]%
        {oldfield2024}
\bibfield{author}{\bibinfo{person}{James Oldfield}, \bibinfo{person}{Markos
  Georgopoulos}, \bibinfo{person}{Grigorios~G Chrysos},
  \bibinfo{person}{Christos Tzelepis}, \bibinfo{person}{Yannis Panagakis},
  \bibinfo{person}{Mihalis~A Nicolaou}, \bibinfo{person}{Jiankang Deng}, {and}
  \bibinfo{person}{Ioannis Patras}.} \bibinfo{year}{2024}\natexlab{}.
\newblock \showarticletitle{Multilinear Mixture of Experts: Scalable Expert
  Specialization through Factorization}.
\newblock \bibinfo{journal}{\emph{arXiv preprint arXiv:2402.12550}}
  (\bibinfo{year}{2024}).
\newblock


\bibitem[Patton et~al\mbox{.}(2016)]%
        {patton_our_2016}
\bibfield{author}{\bibinfo{person}{George~C Patton}, \bibinfo{person}{Susan~M
  Sawyer}, \bibinfo{person}{John~S Santelli}, \bibinfo{person}{David~A Ross},
  \bibinfo{person}{Rima Afifi}, \bibinfo{person}{Nicholas~B Allen},
  \bibinfo{person}{Monika Arora}, \bibinfo{person}{Peter Azzopardi},
  \bibinfo{person}{Wendy Baldwin}, \bibinfo{person}{Christopher Bonell},
  \bibinfo{person}{Ritsuko Kakuma}, \bibinfo{person}{Elissa Kennedy},
  \bibinfo{person}{Jaqueline Mahon}, \bibinfo{person}{Terry McGovern},
  \bibinfo{person}{Ali~H Mokdad}, \bibinfo{person}{Vikram Patel},
  \bibinfo{person}{Suzanne Petroni}, \bibinfo{person}{Nicola Reavley},
  \bibinfo{person}{Kikelomo Taiwo}, \bibinfo{person}{Jane Waldfogel},
  \bibinfo{person}{Dakshitha Wickremarathne}, \bibinfo{person}{Carmen Barroso},
  \bibinfo{person}{Zulfiqar Bhutta}, \bibinfo{person}{Adesegun~O Fatusi},
  \bibinfo{person}{Amitabh Mattoo}, \bibinfo{person}{Judith Diers},
  \bibinfo{person}{Jing Fang}, \bibinfo{person}{Jane Ferguson},
  \bibinfo{person}{Frederick Ssewamala}, {and} \bibinfo{person}{Russell~M
  Viner}.} \bibinfo{year}{2016}\natexlab{}.
\newblock \showarticletitle{Our future: a {Lancet} commission on adolescent
  health and wellbeing}.
\newblock \bibinfo{journal}{\emph{The Lancet}} \bibinfo{volume}{387},
  \bibinfo{number}{10036} (\bibinfo{date}{June} \bibinfo{year}{2016}),
  \bibinfo{pages}{2423--2478}.
\newblock
\showISSN{01406736}
\urldef\tempurl%
\url{https://doi.org/10.1016/S0140-6736(16)00579-1}
\showDOI{\tempurl}


\bibitem[Pearl(2009)]%
        {Pearl2009}
\bibfield{author}{\bibinfo{person}{Judea Pearl}.}
  \bibinfo{year}{2009}\natexlab{}.
\newblock \bibinfo{booktitle}{\emph{Causality: Models, Reasoning and Inference}
  (\bibinfo{edition}{2nd} ed.)}.
\newblock \bibinfo{publisher}{Cambridge University Press}.
\newblock


\bibitem[Polson et~al\mbox{.}(2013)]%
        {Polson2012}
\bibfield{author}{\bibinfo{person}{Nicholas Polson}, \bibinfo{person}{James
  Scott}, {and} \bibinfo{person}{J Windle}.} \bibinfo{year}{2013}\natexlab{}.
\newblock \showarticletitle{Bayesian Inference for Logistic Models Using
  P{\'o}lya--Gamma Latent Variables}.
\newblock \bibinfo{journal}{\emph{J. Amer. Statist. Assoc.}}
  \bibinfo{volume}{108}, \bibinfo{number}{504} (\bibinfo{year}{2013}),
  \bibinfo{pages}{1339--1349}.
\newblock
\urldef\tempurl%
\url{https://doi.org/10.1080/01621459.2013.829001}
\showDOI{\tempurl}
\showeprint{https://doi.org/10.1080/01621459.2013.829001}


\bibitem[Saeed et~al\mbox{.}(2020)]%
        {saeed2020}
\bibfield{author}{\bibinfo{person}{Basil Saeed}, \bibinfo{person}{Snigdha
  Panigrahi}, {and} \bibinfo{person}{Caroline Uhler}.}
  \bibinfo{year}{2020}\natexlab{}.
\newblock \showarticletitle{Causal structure discovery from distributions
  arising from mixtures of dags}. In \bibinfo{booktitle}{\emph{International
  Conference on Machine Learning}}. PMLR, \bibinfo{pages}{8336--8345}.
\newblock


\bibitem[Solmi et~al\mbox{.}(2022)]%
        {solmi_age_2022}
\bibfield{author}{\bibinfo{person}{Marco Solmi}, \bibinfo{person}{Joaquim
  Radua}, \bibinfo{person}{Miriam Olivola}, \bibinfo{person}{Enrico Croce},
  \bibinfo{person}{Livia Soardo}, \bibinfo{person}{Gonzalo Salazar De~Pablo},
  \bibinfo{person}{Jae Il~Shin}, \bibinfo{person}{James~B. Kirkbride},
  \bibinfo{person}{Peter Jones}, \bibinfo{person}{Jae~Han Kim},
  \bibinfo{person}{Jong~Yeob Kim}, \bibinfo{person}{Andr{\`e}~F. Carvalho},
  \bibinfo{person}{Mary~V. Seeman}, \bibinfo{person}{Christoph~U. Correll},
  {and} \bibinfo{person}{Paolo Fusar-Poli}.} \bibinfo{year}{2022}\natexlab{}.
\newblock \showarticletitle{Age at onset of mental disorders worldwide:
  large-scale meta-analysis of 192 epidemiological studies}.
\newblock \bibinfo{journal}{\emph{Molecular Psychiatry}} \bibinfo{volume}{27},
  \bibinfo{number}{1} (\bibinfo{date}{Jan.} \bibinfo{year}{2022}),
  \bibinfo{pages}{281--295}.
\newblock
\showISSN{1359-4184, 1476-5578}
\urldef\tempurl%
\url{https://doi.org/10.1038/s41380-021-01161-7}
\showDOI{\tempurl}


\bibitem[Strobl(2023)]%
        {stobl2023}
\bibfield{author}{\bibinfo{person}{Eric~V. Strobl}.}
  \bibinfo{year}{2023}\natexlab{}.
\newblock \showarticletitle{Causal discovery with a mixture of DAGs}.
\newblock \bibinfo{journal}{\emph{Machine Learning}} \bibinfo{volume}{112},
  \bibinfo{number}{11} (\bibinfo{year}{2023}), \bibinfo{pages}{4201--4225}.
\newblock
\showISBNx{1573-0565}
\urldef\tempurl%
\url{https://doi.org/10.1007/s10994-022-06159-y}
\showDOI{\tempurl}


\bibitem[Tsamardinos et~al\mbox{.}(2006)]%
        {Tsamardinos2006}
\bibfield{author}{\bibinfo{person}{Ioannis Tsamardinos},
  \bibinfo{person}{Laura~E Brown}, {and} \bibinfo{person}{Constantin~F
  Aliferis}.} \bibinfo{year}{2006}\natexlab{}.
\newblock \showarticletitle{The max-min hill-climbing Bayesian network
  structure learning algorithm}.
\newblock \bibinfo{journal}{\emph{Machine Learning}}  \bibinfo{volume}{65}
  (\bibinfo{year}{2006}), \bibinfo{pages}{31--78}.
\newblock
Issue 1.


\bibitem[Varıcı et~al\mbox{.}(2024)]%
        {varıcı2024}
\bibfield{author}{\bibinfo{person}{Burak Varıcı}, \bibinfo{person}{Dmitriy
  Katz-Rogozhnikov}, \bibinfo{person}{Dennis Wei}, \bibinfo{person}{Prasanna
  Sattigeri}, {and} \bibinfo{person}{Ali Tajer}.}
  \bibinfo{year}{2024}\natexlab{}.
\newblock \bibinfo{title}{Interventional Causal Discovery in a Mixture of
  DAGs}.
\newblock
\newblock
\showeprint[arxiv]{2406.08666}~[cs.LG]
\urldef\tempurl%
\url{https://arxiv.org/abs/2406.08666}
\showURL{%
\tempurl}


\bibitem[Vowels et~al\mbox{.}(2023)]%
        {vowels_dya_2023}
\bibfield{author}{\bibinfo{person}{Matthew~J. Vowels},
  \bibinfo{person}{Necati~Cihan Camgoz}, {and} \bibinfo{person}{Richard
  Bowden}.} \bibinfo{year}{2023}\natexlab{}.
\newblock \showarticletitle{D'ya {Like} {DAGs}? {A} {Survey} on {Structure}
  {Learning} and {Causal} {Discovery}}.
\newblock \bibinfo{journal}{\emph{Comput. Surveys}} \bibinfo{volume}{55},
  \bibinfo{number}{4} (\bibinfo{date}{April} \bibinfo{year}{2023}),
  \bibinfo{pages}{1--36}.
\newblock
\showISSN{0360-0300, 1557-7341}
\urldef\tempurl%
\url{https://doi.org/10.1145/3527154}
\showDOI{\tempurl}


\bibitem[Watanabe(2013)]%
        {watanabe2013widely}
\bibfield{author}{\bibinfo{person}{Sumio Watanabe}.}
  \bibinfo{year}{2013}\natexlab{}.
\newblock \showarticletitle{A widely applicable Bayesian information
  criterion}.
\newblock \bibinfo{journal}{\emph{The Journal of Machine Learning Research}}
  \bibinfo{volume}{14}, \bibinfo{number}{1} (\bibinfo{year}{2013}),
  \bibinfo{pages}{867--897}.
\newblock


\end{thebibliography}


\end{document}